\documentclass[sn-basic, Numbered]{sn-jnl}% Basic Springer Nature Reference Style/Chemistry Reference Style
\usepackage[T1]{fontenc}
\usepackage{lmodern}
\usepackage{graphicx}%
\usepackage{multirow}%
\usepackage{amsmath,amssymb,amsfonts}%
\usepackage{amsthm}%
\usepackage{mathrsfs}%
\usepackage[title]{appendix}%
\usepackage{xcolor}%
\usepackage{textcomp}%
\usepackage{manyfoot}%
\usepackage{booktabs}%
\usepackage{algorithm}%
\usepackage{algorithmicx}%
\usepackage{algpseudocode}%
\usepackage{listings}%

\usepackage{multicol}
\usepackage{threeparttable}
\usepackage{subfigure}
\usepackage{hyperref}
\usepackage{mathtools}
\theoremstyle{thmstyleone}%
\theoremstyle{thmstyletwo}%

\theoremstyle{thmstylethree}%

\newcommand{\modelname}{AlignReact}
\begin{document}

\title[Article Title]{%A Foundational Chemical Reaction Representation Learning Framework for Reaction Condition Recommendation and Performance Prediction
A General‑Purpose Framework for Chemical Reaction Representation with Atomic Correspondence and Flexible Condition Adaptation}


\author[1, 2]{\fnm{Kaipeng} \sur{Zeng}}%\email{zengkaipeng@sjtu.edu.cn}
\equalcont{These authors contributed equally to this work.}
\author[1]{\fnm{Xianbin} \sur{Liu}}%\email{zhaoxin0807@sjtu.edu.cn}
\equalcont{These authors contributed equally to this work.}
\author[1]{\fnm{Yu} \sur{Zhang}}%\email{cynthiazhang@sjtu.edu.cn}
\author[1]{\fnm{Xiaokang} \sur{Yang}}%\email{xkyang@sjtu.edu.cn}
\author*[1]{\fnm{Yaohui} \sur{Jin}}\email{jinyh@sjtu.edu.cn}
\author*[1]{\fnm{Yanyan} \sur{Xu}}\email{yanyanxu@sjtu.edu.cn}
\affil*[1]{\orgdiv{MoE Key Laboratory of Artificial Intelligence, AI Institute}, \orgname{Shanghai Jiao Tong University}, \orgaddress{\city{Shanghai}, \postcode{200240}, \state{Shanghai}, \country{China}}}

%\affil[2]{\orgdiv{Data-Driven Management Decision Making Laboratory}, \orgname{Shanghai Jiao Tong University}, \orgaddress{\street{Dongchuan Road}, \city{Shanghai}, \postcode{200240}, \state{Shanghai}, \country{China}}}
\affil[2]{\orgname{Shanghai Innovation Institute}, \orgaddress{\city{Shanghai}, \postcode{200231}, \state{Shanghai}, \country{China}}}


\abstract{
% \textbf{Motivation} Organic synthesis is fundamental to the chemical industry, especially in fields like pharmaceutical development. While machine learning offers great potential for enhancing tasks related to organic reactions, current approaches often depend on hand-crafted features or directly borrow architectures from other domains. These methods struggle to scale effectively with increasing data volumes and frequently fail to capture the nuanced chemical information intrinsic to reaction processes. %To bridge this gap, we aim to develop a robust and scalable representation learning framework that explicitly models molecular transformations and adapts flexibly to diverse reaction conditions and downstream applications.
% \textbf{Motivation} {Organic synthesis is fundamental to the chemical industry, particularly in domains such as pharmaceutical development. While artificial intelligence offers powerful tools for modeling chemical reactions, current approaches are primarily limited to two paradigms: those that rely on hand-crafted, domain-specific features, and those that apply generic deep learning models through simplistic concatenation or aggregation of reaction components. The former often struggles to scale effectively with increasing data volumes, while the latter fails to capture and distinguish the diverse, chemically meaningful relationships between different components within a reaction system. This fundamental limitation hinders the effective utilization of rich chemical information inherent in reaction processes.}
\textbf{Motivation} Organic synthesis is fundamental to the chemical industry, particularly in domains such as pharmaceutical development. While artificial intelligence offers powerful tools for modeling chemical reactions, current approaches are primarily limited to two paradigms: those that rely on hand-crafted, domain-specific features, and those that apply generic deep learning models through simplistic concatenation or aggregation of reaction components. The former often struggles to scale effectively with increasing data volumes, while the latter relies on simple input- or feature-level concatenation to combine different reaction components. Such simplistic aggregation prevents these models from directly capturing the structural transformations between reactants and products, and also makes it difficult to adapt or extend them to datasets that include non-molecular reaction conditions without modifying their model architectures.
\vspace{2mm}\\
\textbf{Results} This paper introduces {\modelname}, a novel chemical reaction representation learning framework designed for a wide range of organic reaction tasks. Our approach integrates atomic correspondence between reactants and products to discern precise molecular transformations, thereby improving the model's comprehension of molecular transformation patterns. We incorporate an adapter structure to embed reaction conditions into the representation, enhancing adaptability across varied datasets and tasks. Furthermore, a Reaction-Center-Aware attention mechanism is proposed to enable the model to focus on critical functional groups, yielding more powerful and informative representations. Evaluated across multiple downstream tasks, our model demonstrates superior performance, significantly outperforming existing chemical reaction representation learning architectures on most benchmark datasets.
\vspace{2mm}\\
\textbf{Scientific contribution} We introduce a chemical reaction representation learning framework that explicitly integrates atomic correspondence between reactants and products into the network architecture, enabling the model to perceive and model molecular structural transformations during reactions. %Our approach also features a flexible, detachable module for integrating reaction conditions, which accepts precomputed features of varying dimensions and modalities, offering greater flexibility and broader dataset compatibility compared to prior work, and enables fine-grained, context-aware conditioning. Extensive experiments on a range of downstream datasets demonstrate that our framework achieves state-of-the-art performance across several key chemical reaction prediction tasks.
{As an extensible but preliminary feature, our approach also features a flexible, detachable module for integrating reaction conditions. Inspired by conditioning mechanisms from multimodal generation, this adapter module accepts precomputed features of varying dimensions and modalities, laying the groundwork for broader dataset compatibility compared to prior work, and enables fine-grained, context-aware conditioning.} Extensive experiments on a range of downstream datasets demonstrate that our framework achieves state-of-the-art performance across several key chemical reaction prediction tasks.}




\maketitle

\section{Introduction}\label{sec1}

Molecular design and synthesis constitute the core processes within the organic chemical industry~\cite{dara2022machine,hasselgren2024artificial,ball2015chemistry,mikulak2020computational}. For instance, in drug discovery, synthesizing and designing molecules with specific properties facilitates the creation of therapeutic candidates targeting various diseases~\cite{shen2021automation,zhang2025large}. Despite significant advancements in chemical instrumentation over recent decades, the design-synthesis cycle remains highly time-consuming~\cite{angello2022closed,burger2020mobile,rinehart2023machine}. This complexity arises not only from navigating the vast chemical reaction space for molecular design and synthesis pathway planning but also from selecting appropriate reagents and identifying optimal reaction conditions.

Given this inherent complexity, researchers have increasingly turned to artificial intelligence (AI) for assistance. The surge in available data, coupled with advances in AI, has enabled substantial progress in integrating these technologies into organic chemistry~\cite{sanchez2018inverse,wang2023scientific,zeng2024ualign}. However, existing studies primarily focus on molecular design tasks, such as virtual screening~\cite{zhou2023unimol,yang2022learning}, which typically involve modeling individual molecules. In contrast, tasks crucial for guiding chemical synthesis processes, such as reaction condition recommendation or reaction selectivity prediction, require modeling the interactions between various components within the reaction system. These synthesis-oriented tasks remain relatively unexplored and continue to pose significant challenges. Effectively addressing them hinges on developing robust chemical reaction representations capable of capturing complex molecular relationships during reactions. Therefore, this work focuses on designing a backbone architecture for chemical reaction representation learning, aiming to advance the state-of-the-art in this area.

Current methods for chemical reaction representation learning can be broadly categorized into two groups: fingerprint-based and deep learning-based. Fingerprint-based methods~\cite{probst2022reaction,MFF_SANDFORT20201379,Gao2018} utilize handcrafted molecular representations (fingerprints) and employ various strategies to integrate the representations of different reaction components into a unified reaction representation. These representations are often used with conventional machine learning models such as XGBoost~\cite{chen2016xgboost}. While computationally efficient and effective with smaller datasets, these approaches may exhibit performance limitations when scaled to larger datasets and more complex scenarios due to the inherent oversimplification of information in their manually designed, statistics-based features~\cite{ji2023drugood}.

To pursue more powerful representations, researchers have increasingly adopted deep learning techniques. The use of textual representations of molecules, such as SMILES ~\cite{weininger1988smiles}, enables molecular encoding with language models~\cite{irwin2022chemformer,schwaller2019molecular}. Building upon this, extensions have been developed that combine textual representations of different reaction components for chemical reaction representation learning using language models~\cite{lu2022unified,schwaller2021mapping}. However, such sequential representations are highly dependent on the traversal order of molecular graphs and may overlook molecular topological information~\cite{yang2022learning}. This limitation can be amplified in reaction representation learning involving multiple molecules. Beyond linear encoding, structure-based methods have also emerged. Given the inherent graph structure of molecules, researchers turn to leveraging graph neural networks (GNNs) for chemical reaction representation learning~\cite{maser2021multilabel,han2024improving,chemprop:doi:10.1021/acs.jcim.3c01250,van20243dreact,xu2025unified,wang2025reacon}. Compared to single-molecule representation learning, some studies incorporate domain-specific adaptations. For instance, in yield prediction, where data is often noisy, studies~\cite{chen2024uncertainty,kwon2022uncertainty} adjust the training loss to incorporate uncertainty, thereby refining model performance. Other works~\cite{dataset2_denmark_stereo_zahrt2019prediction,dataset2_hongxin_nc_stereo_li2023reaction,dataset1_hongxin_Angew_regio_li2020predicting,coley_qm_gnn_guan2021regio} integrate DFT-derived descriptors of reaction-influencing factors as enhancements. Nevertheless, these deep learning methods typically either directly apply frameworks from other domains or independently extract features for each reaction component before utilizing simplistic aggregation mechanisms, such as concatenation or summation, to derive the reaction representation. % These approaches still fail to model the interactions between different reaction components adequately, overlooking the rich contextual information inherent in complex chemical reactions.
While straightforward, this simplistic modeling approach overlooks a fundamental aspect of reaction data: the structural transformation from reactants to products. By simply concatenating input data or applying a simple feature aggregation after processing reactants and products independently, these methods lack an explicit mechanism to perceive how atoms are conserved and how bonds are rearranged during a reaction. While it is possible for sufficiently expressive models to implicitly learn such patterns from large amounts of data, this inductive bias is not built into their architecture, making the learning process less efficient and the resulting representations potentially less powerful. This limitation hinders the model's ability to fully capture the rich, domain-specific relational information inherent in chemical reactions, ultimately constraining its performance across a range of downstream tasks.

Beyond under-utilizing reaction information, current methods often inadequately handle reaction conditions. Existing approaches typically focus exclusively on the molecular components of those conditions, such as reagents. Moreover, these methods either simplistically treat reaction conditions as part of the reactants or employ fixed sub-network architectures for processing them. The former approach fails to differentiate the functional roles of distinct reaction components and overlooks their interactions. The latter lacks flexibility, hindering adaptation to diverse datasets with varying component annotations across different tasks. Critically, such limitations prevent effective applications to datasets incorporating non-molecular conditions that  significantly influence chemical reactions, such as temperature~\cite{Wang2023,doi:10.1021/ci900437n}, or scenarios requiring interpretation of detailed natural language experimental procedures~\cite{kearnes2021open}. Consequently, there is a need for a chemical reaction representation learning model that can integrate richer chemical information and flexibly accommodate diverse modalities of reaction conditions.

To address these limitations, we propose {\modelname}, a novel chemical reaction representation learning framework for multiple downstream tasks. Recognizing that reaction centers and the specific bond changes involved are pivotal in determining reaction outcomes~\cite{schwaller2021mapping}, we draw inspiration from the concept of imaginary transition structures in organic reactions~\cite{fujita1986description}. Our model features an Atom-Aligned Encoder that explicitly captures structural changes by enabling information fusion between corresponding atom pairs in the reactants and products. Additionally, we introduce a Reaction-Center-Aware Decoder to help the model focus on key functional groups involved in the transformation. {To flexibly integrate diverse reaction conditions (molecular and non-molecular), we employ an Adapter module as an extensible yet preliminary feature. Drawing architectural inspiration from multimodal generation models such as T2I-Adapter~\cite{mou2024t2i} and ControlNet~\cite{zhang2023adding}, this design serves as a foundational step to enhance our model's adaptability to datasets with varying types and numbers of reaction condition components.} We evaluated {\modelname} on various tasks, including reaction condition prediction, reaction yield prediction, and reaction selectivity prediction. Our model demonstrates significant performance improvements across all tasks, outperforming the baselines, even those leveraging extensive pretraining.

\section{Results} \label{sec2}
\subsection{\modelname}

\begin{figure}
    \centering
    \includegraphics[width=\linewidth]{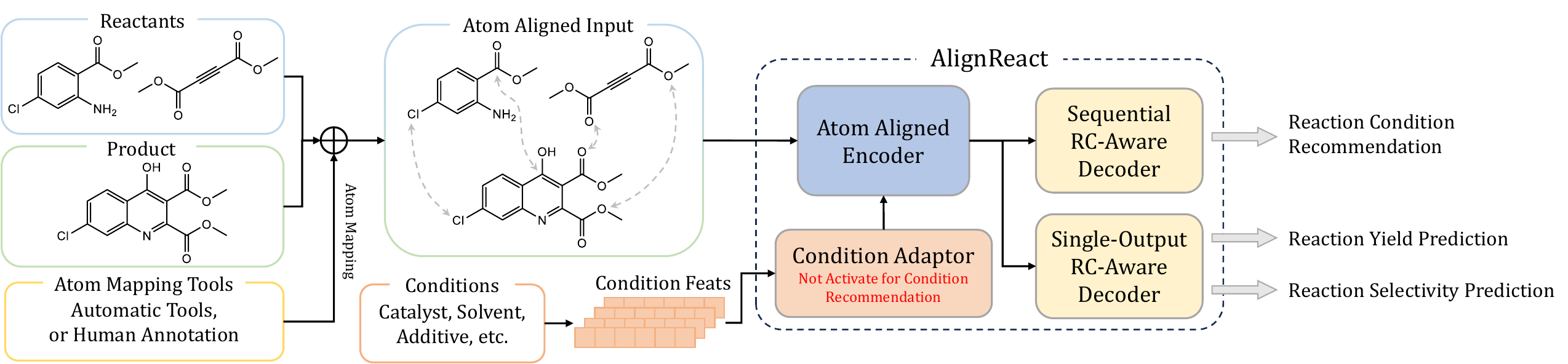}
    \caption{\textbf{Overview.} Illustration of {\modelname}'s pipeline and its functionality: reaction condition recommendation, yield prediction, and selectivity prediction. }
    \label{fig: pipeline}
\end{figure}

We present \textbf{AlignReact}, a general-purpose framework for learning chemical reaction embeddings that adapt to diverse, reaction-conditioned tasks. As illustrated in Fig.~\ref{fig: pipeline}, AlignReact supports two distinct task paradigms: (1) \textit{variable-length output tasks}, where the target label contains an arbitrary number of components (e.g., a set of reagents, catalysts, and solvents); and (2) \textit{single-output tasks}, which require a scalar or categorical prediction (e.g., reaction yield or selectivity).  

{To extract powerful reaction representations that are adaptable to these heterogeneous tasks, our architecture centers on two primary innovations: the Atom-Aligned Encoder and the RC-Aware Decoder, supplemented by an extensible Condition Adapter. Each component is designed to incorporate specific chemical priors and maximize flexibility.}

\begin{itemize}
    \item \textbf{Atom-Aligned Encoder.}  
    This component processes the molecular graphs of reactants and products (atoms as nodes, bonds as edges) through a series of graph neural network layers. After each message-passing step, an \textit{information fusion layer} exchanges features between corresponding atom pairs in the reactants and products, guided by the provided atom-mapping. This forced alignment enables the model to directly capture bond changes and atom conservation patterns, i.e., the precise structural transformations that define a chemical reaction. By explicitly modeling these changes rather than treating reactants and products independently, the encoder learns a representation that is sensitive to the structural evolution from reactants to products, which is fundamental for downstream tasks.

    \item \textbf{Condition Adapter.}  
    Reaction conditions (e.g., reagents, temperature, solvents, or textual procedure descriptions) vary widely across datasets and modalities. { Presented here as an extensible preliminary feature,} the Condition Adapter is a flexible module that injects condition information into the encoder’s node representations. It accepts precomputed condition features encoded by any modality-specific encoder (e.g., a graph neural network for molecules, a multilayer perceptron for scalars, or a pretrained language model for text). At every encoder block, these condition features are integrated via multi-head cross-attention, followed by a residual connection that preserves the original node features. This design is \textit{detachable}; the adapter can be included or omitted depending on data availability, and it naturally accommodates an arbitrary number and type of condition entries. This offers greater flexibility and compatibility compared to previous reaction representation models, which typically treat conditions as part of the reactants or rely on fixed, non-adaptive subnetworks.

   \item \textbf{Reaction-Center-Aware (RC-aware) Decoder.}  The decoder takes the final node embeddings from the Atom-Aligned Encoder and produces task-specific outputs. Its core is an RC-aware cross-attention mechanism that explicitly incorporates knowledge of the reaction center—the atoms directly involved in bond changes, their neighbors, and leaving groups. Part of the attention heads are constrained to attend only to reaction-center atoms, while the remaining heads perform global attention over all nodes. This design encodes a chemical prior: by focusing a subset of heads on the reaction center, the model is relieved from learning to identify the transformation site from scratch. This allows the model to allocate its capacity more effectively to peripheral functional groups that may influence the chemical reaction while ensuring that the core transformation is always attended to. The decoder is instantiated in two variants to cover the two task paradigms:
\begin{itemize}
    \item For \textbf{single-output tasks} (e.g., yield or selectivity prediction), a \textit{learnable query vector} is introduced. This vector serves as the query in the RC-aware cross-attention, effectively pooling the node embeddings into a fixed-dimensional reaction representation, which is then passed through a multilayer perceptron to obtain the final prediction.
    \item For \textbf{variable-length output tasks} (e.g., reaction condition recommendation), the decoder operates autoregressively. It generates the output sequence item by item; at each step, the previously generated items (the \textit{sequential input}) are fed into the decoder along with the node features from the encoder, enabling the model to predict the next item.
\end{itemize}
\end{itemize}

\begin{figure}
    \centering
    \includegraphics[width=\linewidth]{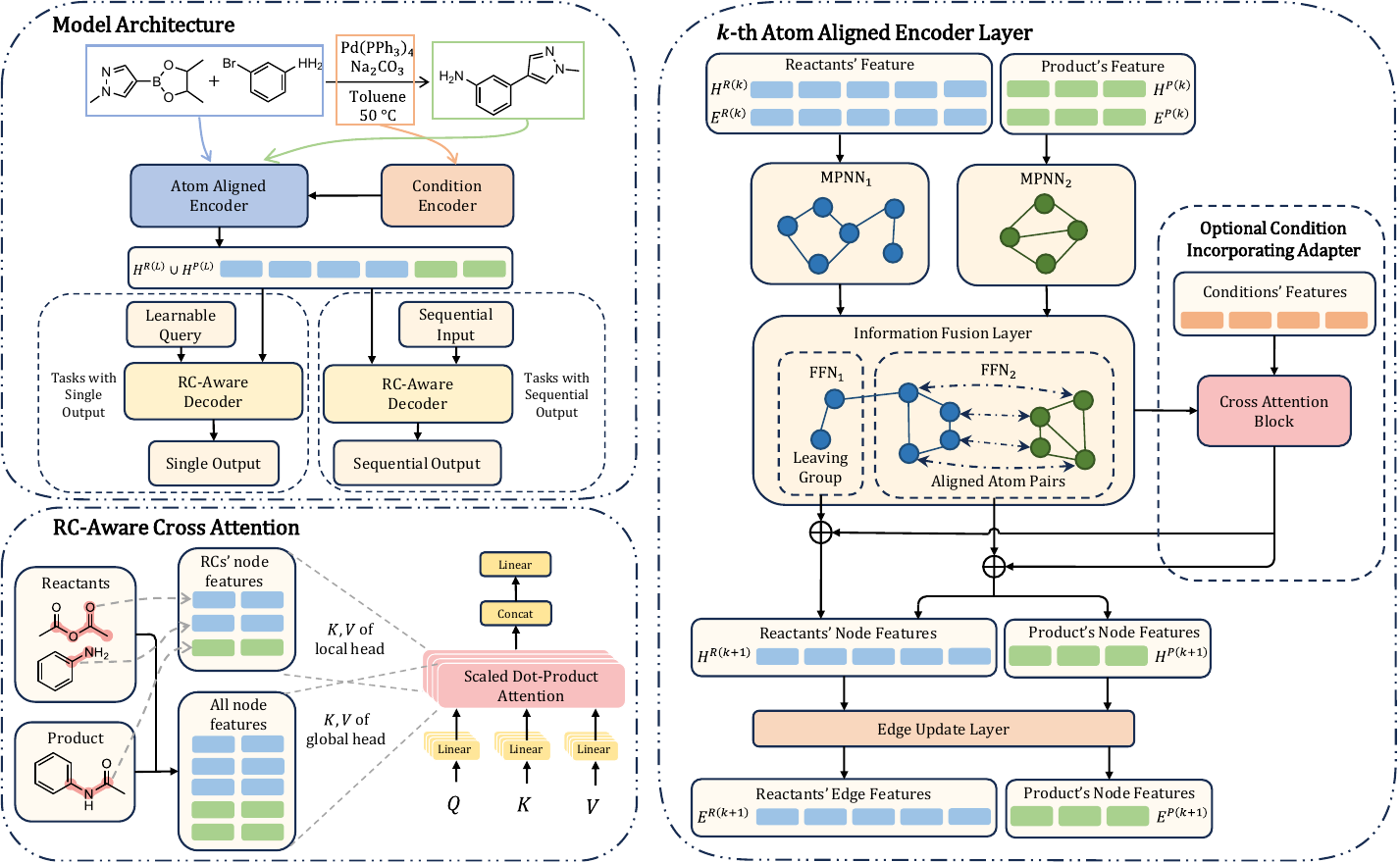}
    \caption{\textbf{Model Architecture of {\modelname}}. For a given chemical reaction, the molecule graphs of reactants and products, along with their atomic correspondence, are input into a $L$-layer Atom-Aligned Encoder to extract reaction node features $H^{P(L)}$ and $H^{R(L)}$. If the reaction conditions are also provided, they are encoded by a condition encoder and merged into $H^{P(L)}$ and $H^{R(L)}$ via an adapter structure  proposed in this study. The resulting features are then processed by the RC-aware decoder to produce outputs for subsequent tasks.  We have tailored two decoders for both sequential output and single output, both featuring an RC-aware cross-attention layer to concentrate on key reaction motifs.
    }
    \label{fig: overall}
\end{figure}

All components are visualized in Fig.~\ref{fig: overall}, and comprehensive implementation details are provided in Sec.~\ref{methods} and Supplementary Sec.~\ref{impdetail}. By explicitly separating the roles of atom alignment, condition integration, and reaction-center awareness, AlignReact offers a transparent and adaptable architecture that outperforms existing methods across a range of reaction-prediction tasks.

\subsection{Performance of {\modelname} on open benchmark data}
To demonstrate that \modelname's chemical reaction representations effectively handle diverse reaction-related tasks, we benchmarked the model across multiple downstream applications. These include reaction condition recommendation, yield prediction, and selectivity prediction (encompassing both regio-selectivity and enantio-selectivity), as previously mentioned. Notably, due to variations in dataset sizes and input characteristics across tasks, we trained specialized models for each downstream dataset.  Since our model requires atom mapping as input, we employed RXNMapper~\cite{schwaller2021extraction} to automatically annotate atom mappings for all datasets except the Buchwald–Hartwig cross‑coupling reaction dataset, which is used for the reaction yield prediction task. This process was fully automatic and involved no manual corrections. Both training and inference were performed on this automatically annotated data. For the Buchwald–Hartwig dataset, which follows a single reaction template, we adopted a rule‑based approach for atom‑mapping annotation. Detailed data processing procedures are provided in Supplementary Sec.~\ref{app: datapre}. Experimental results are presented in Fig.~\ref{fig:reaction condition}, Fig.~\ref{fig: cnresuts}, and Fig.~\ref{fig: sel-results}. The full experiment results are provided in Supplementary Sec.~\ref{sec:si_full_results}.
\subsubsection{Reaction Condition Recommendation}
\begin{figure}
    \centering
    \includegraphics[width=\linewidth]{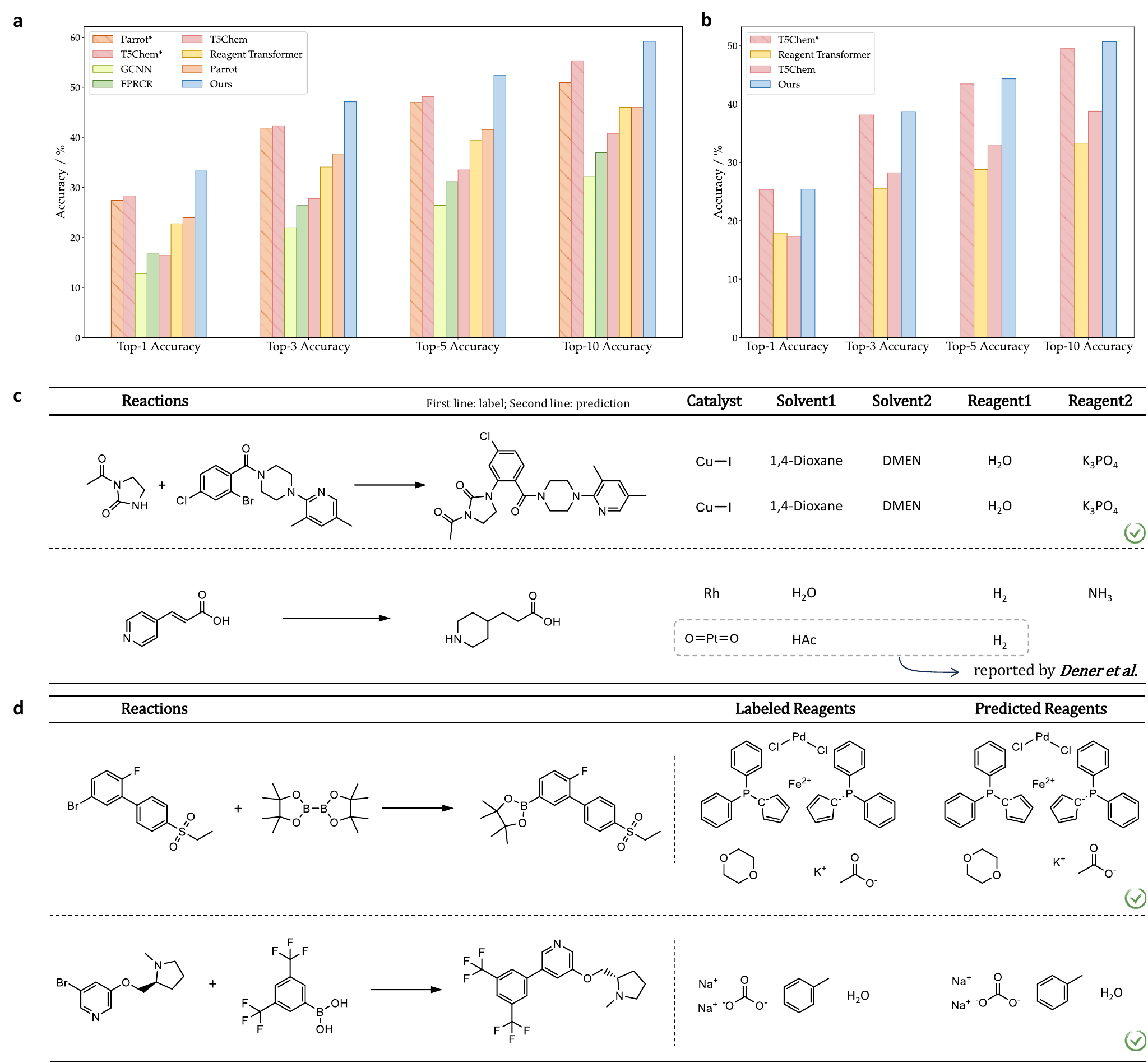}
    \caption{\textbf{The performance of different models for reaction condition recommendation.} \textbf{a,} The performance of different models on USPTO-Condition dataset, *denotes the model is pretrained on large-scale reaction dataset.  \textbf{b,} The performance of different models on USPTO-500MT dataset, *denotes the model is pretrained on large-scale reaction dataset. \textbf{c,} Visualization of generated conditions on USPTO-Condition dataset, one C-N cross-coupling reaction and  one pyridine reduction reaction are selected.  Each row represents a single example. In each row, the first line shows the ground truth (label), and the second line shows the model’s prediction. \textbf{d,} Visualization of generated condition of two out-of-distribution samples.}
    \label{fig:reaction condition}
\end{figure}
We evaluate our model on reaction condition recommendation, where conditions refer to chemical reagents beyond the core reactants. This task is formulated as molecular sequence generation. Experiments are conducted on two datasets: USPTO-Condition and USPTO-500MT. For the USPTO-Condition, we follow the data processing and splits from Parrot-LM-E~\cite{Wang2023}, comprising 544,591 training samples, 68,075 validation samples, and 68,075 test samples. Three SMILES-based baselines~\cite{Wang2023,lu2022unified,andronov2023reagent} are selected for comparison. Given that the reaction conditions in this dataset consistently comprise one catalyst, two solvents, and two reagents, we adapt two predictive baselines~\cite{Gao2018,maser2021multilabel} accordingly. For USPTO-500MT, we adopt the data partitioning from T5Chem~\cite{lu2022unified}, which contains 116,360 training samples, 12,937 validation samples, and 14,238 test samples. The strategy of molecular ordering in labels is presented in Supplementary Sec.~\ref{app: pre_500_mt}. The same three SMILES-based baselines used for USPTO-Condition are employed for comparison. For both datasets, we use top-$k$ accuracy as the evaluation metric, defined as the proportion of test cases in which the correct answer appears among the top-$k$ candidates. A candidate is considered correct if the predicted molecules exactly match the label.

As shown in Fig.~\ref{fig:reaction condition}a, our model achieves 33.30\% top-1 accuracy, 52.43\% top-5 accuracy, and 59.19\% top-10 accuracy on the USPTO-Condition dataset. This represents absolute improvements of 10.56\% in top-1 accuracy, 13.07\% in top-5 accuracy, and 13.18\% in top-10 accuracy over the strongest non-pretrained baseline, Reagent Transformer. Notably, our approach also exceeds the top pretrained baseline, T5Chem, by 4.99\% in top-1 accuracy, 4.28\% in top-5 accuracy, and 3.87\% in top-10 accuracy. From Fig.~\ref{fig:reaction condition}b, on the USPTO-500MT dataset, our model achieves a top-1 overall accuracy of 25.39\% and a top-10 overall accuracy of 50.65\%, exceeding the strongest baseline that did not utilize pretraining by 9.03\% and 11.93\%. It can be observed that our model also outperforms the T5Chem model pretrained on a large-scale reaction dataset.  The aforementioned performance indicates that our model is capable of extracting robust reaction representations for reaction condition prediction, even without the aid of large-scale pretraining.

To complement the macro-scale evaluation metrics and provide a more concrete understanding of the model's performance, we conducted a series of illustrative case studies, as summarized in Fig.~\ref{fig:reaction condition}c and d. As shown in Fig.~\ref{fig:reaction condition}c, our model accurately predicted the reagents required for a C–N cross-coupling reaction. We also examined a pyridine reduction reaction from the test set. Although the model failed to reproduce the exact ground-truth conditions, it instead suggested an alternative set of conditions reported to be high-yielding in a patent~\cite{condition_pattern1}. Fig.~\ref{fig:reaction condition}d presents the model’s predictions for two out-of-distribution examples~\cite{doi:10.1021/acs.jmedchem.9b00322,lin20023}, for which the model successfully predicted all required reagents. These results demonstrate that the model is not only capable of accurately recalling established knowledge from its training domain but also of intelligently recommending chemically intuitive reagents, indicating a capacity for extrapolation and generalization.

\subsubsection{Yield Prediction}
We evaluate our model on the high-throughput experimentation (HTE) yield dataset for Buchwald-Hartwig cross-coupling reactions~\cite{ahneman2018predicting}, which serves as a standard benchmark for catalytic reaction prediction and captures the complexity of palladium-catalyzed C–N bond formation. Following YieldBert~\cite{schwaller2021prediction}, we adopt 10 random splits and 4 out-of-distribution (OOD) splits for evaluation. Our baselines include two non-deep learning methods~\cite{probst2022reaction,MFF_SANDFORT20201379} that combine handcrafted features with XGBoost~\cite{chen2016xgboost}, as well as four deep learning approaches: two SMILES-based models~\cite{schwaller2021prediction,lu2022unified}, one graph-based model~\cite{chemprop:doi:10.1021/acs.jcim.3c01250}, and one multi-modal model~\cite{shi2024prediction}. A Graph Attention Network (GAT)~\cite{velivckovic2018gat} serves as our condition encoder. Given the limited chemical diversity of the dataset and the availability of pretrained baselines, we further implement a variant that leverages a lightweight pretrained molecular encoder~\cite{Hu*2020Strategies} for condition representation. Implementation details are provided in Supplementary Sec.~\ref{app: condenc}.

Fig.~\ref{fig: cnresuts} summarizes the experimental results, including representative scatter plots of predicted versus experimental yields under a random split and an OOD split. As shown in Fig.~\ref{fig: cnresuts}d, our model equipped with a pretrained condition encoder achieves an average RMSE of 5.0662 across ten random splits, outperforming all pretrained baselines by a substantial margin. Notably, even without a pretrained condition encoder, our model still surpasses large-scale pretrained baselines such as T5Chem under random splits. Both variants of our model also exhibit strong OOD generalization. For instance, without pretrained condition encoders, our model attains an average RMSE of 11.3108, substantially outperforming all non-pretrained baselines. When enhanced with a pretrained condition encoder, our model outperforms all baselines on every OOD split except Test4. These results highlight the architectural advantages of our framework. Furthermore, the consistent improvements gained by integrating pretrained encoders demonstrate that our adapter design can seamlessly incorporate external knowledge or pre-trained representations, underscoring its architectural flexibility.

\begin{figure}
    \centering
    \includegraphics[width=\linewidth]{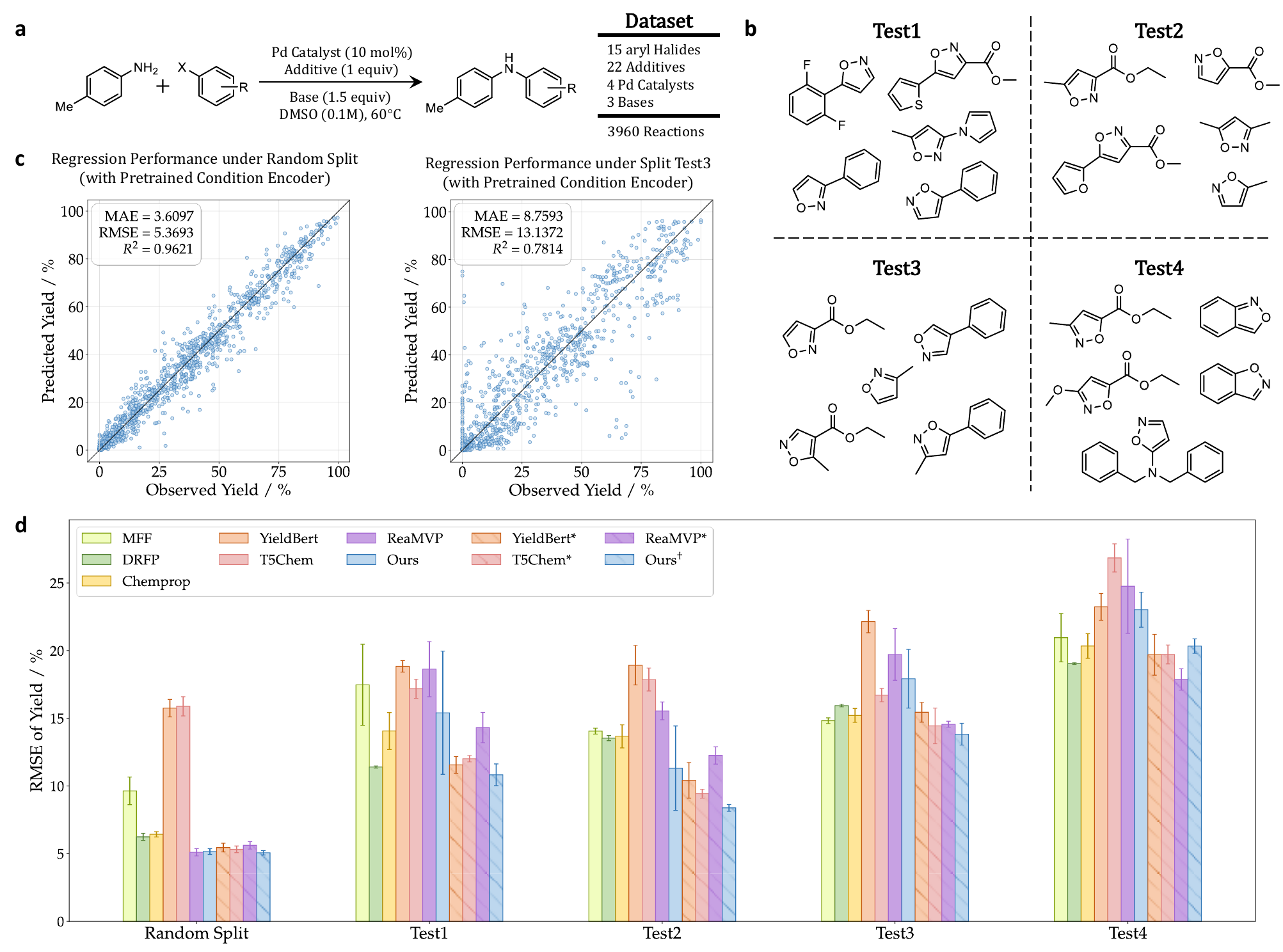}
    \caption{\textbf{The performance of different models on yield prediction.} Full experimental results are presented in Supplementary Sec.~\ref{sec:si_yield_results}. \textbf{a,} an overview of Buchwald-Hartwig cross-coupling reaction dataset. \textbf{b,} Illustration of the four out-of-distribution (OOD) splits in the Buchwald–Hartwig cross-coupling reaction dataset. The splits are constructed based on additives in the reaction data. Additives assigned to each Test\emph{i} subset appear exclusively in that test set, while the remaining additives are used in the training and validation sets. \textbf{c,} Representative scatter plots of predicted versus experimental values under both a random split and the OOD split Test3. \textbf{d,} Model performance comparison under different splits of the Buchwald–Hartwig cross-coupling reaction dataset. * denotes models pretrained on large-scale reaction datasets; $^\dag$ denotes models equipped with pretrained condition encoders. Error bars indicate standard deviation.}
    \label{fig: cnresuts}
\end{figure}

\subsubsection{Selectivity Prediction}
\begin{figure}
    \centering
    \includegraphics[width=\linewidth]{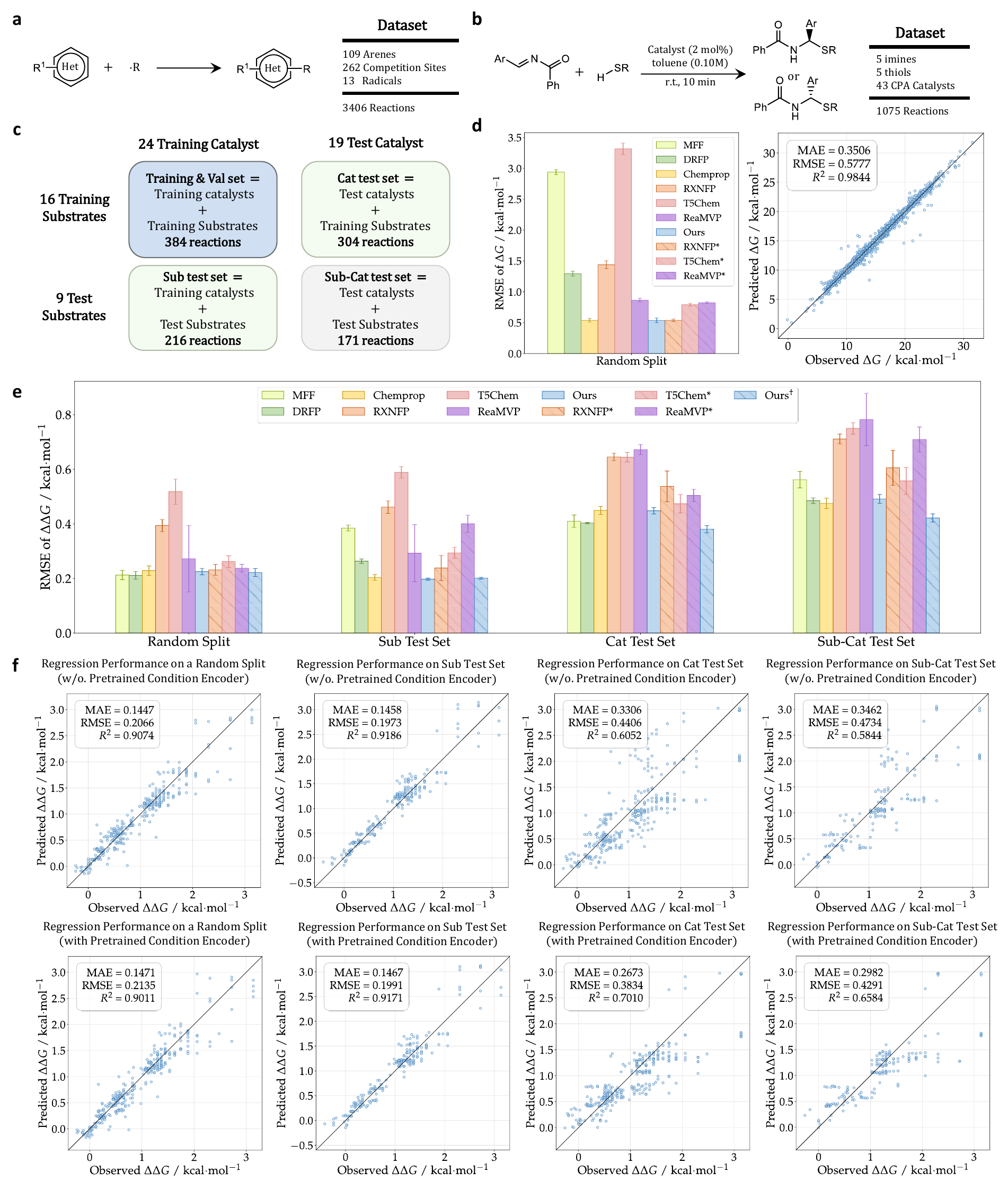}
    \caption{\textbf{Performance of different models on selectivity prediction.} Full experimental details are provided in Supplementary Sec.~\ref{sec:si_selectivity_results}. In panels \textbf{d} and \textbf{e}, * denotes models pretrained on large-scale reaction datasets, and $^\dag$ denotes models equipped with pretrained condition encoders. Error bars represent standard deviation. \textbf{a,} Overview of the radical C–H functionalization dataset. \textbf{b,} Overview of the chiral phosphoric acid-catalyzed thiol addition dataset. \textbf{c,} Out-of-distribution (OOD) splitting strategy for the thiol addition dataset. \textbf{d,} Model performance under random split of the radical C–H functionalization dataset: bar plot comparison of all models and representative scatter plot of our model's predicted vs. experimental values. \textbf{e,} Model performance comparison under random splits and three OOD test sets of the thiol addition dataset. \textbf{f,} Representative scatter plots of our model's predicted vs. experimental values under random split and three OOD test sets of the thiol addition dataset.}
    \label{fig: sel-results}
\end{figure}
We evaluate our model on two benchmark datasets for regio- and enantioselectivity prediction. 
\begin{itemize}
    \item[1)] \textbf{Regioselectivity.} We use the radical C–H functionalization dataset~\cite{dataset1_hongxin_Angew_regio_li2020predicting}, which comprises reactions spanning diverse heterocyclic arene scaffolds, substitution patterns, and radical species. Each entry provides SMILES strings for the substrate and the product, along with computed DFT free energy barriers ($\Delta G$) as quantitative selectivity metrics.
    \item[2)] \textbf{Enantioselectivity.} We adopt the chiral phosphoric acid-catalyzed thiol addition dataset~\cite{dataset2_denmark_stereo_zahrt2019prediction,dataset2_hongxin_nc_stereo_li2023reaction}. Each entry consists of a reactant pair, a product, and a chiral catalyst, with the label determined by the difference in free energy barrier ($\Delta \Delta G$) between the two enantiomeric pathways. For model input, we use the product molecular graphs with all chiral identifiers removed.
\end{itemize}
For both datasets, we perform ten independent random splits, reserving 30\% of the data as the test set to ensure rigorous evaluation. The chiral phosphoric acid-catalyzed thiol addition dataset was also partitioned into training and test sets, following the out-of-distribution (OOD) split based on substrates and catalysts, as defined in the original work~\cite{dataset2_denmark_stereo_zahrt2019prediction}. From the training set, 10\% of the data was further reserved as a validation set for checkpoint selection. Five random seeds are used for model training under the OOD split.

Given that reaction selectivity prediction is inherently a regression task, we adapt three state-of-the-art deep learning models originally designed for reaction yield prediction~\cite{chemprop:doi:10.1021/acs.jcim.3c01250,lu2022unified,schwaller2021mapping,shi2024prediction}. We also compare our model against two fingerprint-based approaches~\cite{probst2022reaction,MFF_SANDFORT20201379}. For the thiol addition dataset, which features catalysts as reaction conditions, we followed the protocol from our yield prediction experiments by implementing two model variants: one with a simple GAT condition encoder and another utilizing a pretrained molecular encoder.

Fig.~\ref{fig: sel-results} summarizes the experimental results for two datasets; full details are available in Supplementary Sec.~\ref{sec:si_selectivity_results}. As shown in Fig.~\ref{fig: sel-results}d, our model achieves state-of-the-art performance on the C–H functionalization dataset with an average RMSE of 0.5375 kcal$\cdot$mol$^{-1}$, surpassing even pretrained models such as ReaMVP and T5Chem. These results confirm that our architectural innovations yield a better model of molecular transformations in chemical reactions, especially in contexts limited to inter-reactant interactions.

The thiol addition dataset presents a more challenging learning scenario due to its limited chemical space (only 10 reactants and 43 catalysts), which inherently restricts the efficacy of deep learning models. As shown in Fig.~\ref{fig: sel-results}e, under random split, fingerprint-based methods deliver the best results; nevertheless, both variants of our model achieve performance on par with these top performers while outperforming all other deep learning baselines. For instance, the variant without pretrained condition encoders attains an average RMSE of 0.2215 kcal$\cdot$mol$^{-1}$, which is the closest to DRFP's 0.2109 kcal$\cdot$mol$^{-1}$ among all deep learning models. Under the more stringent out-of-distribution (OOD) splits, fingerprint-based methods maintain an edge; yet our model still surpasses all other deep learning architectures and remains competitive with fingerprint-based methods. Remarkably, when equipped with a pretrained condition encoder, our model even exceeds the fingerprint-based baseline on the two catalyst-based OOD splits. These findings collectively demonstrate that \modelname\ exhibits strong robustness under both data scarcity and distribution shifts.

\section{Discussion}\label{sec12}

In this paper, we propose {\modelname}, a novel chemical reaction representation learning model. Our model integrates the atomic correspondence between reactants and products, as well as information about the reaction center, enabling it to better model the reaction-induced structural changes and gain a deeper understanding of the  molecular transformation patterns. An adapter is utilized to incorporate reaction conditions, allowing our model to adapt to various modalities of reaction conditions and efficiently leverage previous work to enhance performance. Experimental results demonstrate that our model architecture outperforms existing reaction representation learning architectures across various downstream tasks.  Future work will focus on scaling up pretraining to harness the reaction representation space as a tool for mapping reactivity patterns, thereby assisting chemists in rational reaction design and hypothesis generation.

\textbf{Limitations.} \modelname\ requires atom mappings as input. We emphasize that with the aid of contemporary automated atom-mapping tools, the model achieves competitive performance without the need for manual annotation. Nevertheless, while infrequent with current tools, atom-mapping errors can still negatively impact model predictions. For applications demanding exceptionally high precision, human verification and correction of atom mappings remain indispensable. Like most deep learning methods, our model requires a substantial amount of training data for support; therefore, on small-scale datasets, our model cannot surpass handcrafted features.

\section{Methods}\label{methods}

% We introduce a novel chemical reaction feature extractor named {\modelname} with an encoder-decoder architecture, as illustrated in Fig.~\ref{fig: overall}. The encoder incorporates the atomic correspondence between reactants and products to generate robust features for the chemical reaction. The decoder then integrates the output features from the encoder according to the reaction center information to produce differently formatted outputs customized to downstream tasks.

\subsection{Reaction Graphs}
\modelname takes a pair of molecular graphs as input, representing the reactants and products, respectively. We denote the reactant graph as $G_R=(V_R, E_R)$ and the product graph as $G_P = (V_P, E_P)$, where $V$ (vertices) represent atoms and $E$ (edges) represent chemical bonds. 

According to the principle of atom conservation in chemical reactions, each atom in the products corresponds uniquely to an atom in the reactants. Let $V_R = \{v^R_{1}, v^R_{2}, \ldots, v^R_{n}\}$ and $V_P = \{v^P_{1}, v^P_{2}, \ldots, v^P_{m}\}$ with $n \ge m$. For clarity and consistency in subsequent discussions, we assume that for all $1\le i \le m$, the reactant atom $v^P_{i}$ uniquely corresponds to the product atom $v^R_{i}$. Atoms present in the reactants but not in the products are defined as the leaving group, denoted as $V_L=\{v_{m+1}^R,v_{m+2}^R,\ldots,v_{n}^R\}$.

We further define the set of reaction centers, denoted \(V_{rc}\), as a subset of atoms from both reactants and products. This definition extends conventional reaction center concepts and includes an atom if it meets any of the following criteria:
\begin{itemize}
    \item It is a terminal atom of any chemical bond that changes during the reaction.
    \item Its hydrogen atom count differs from that of its counterpart in the opposing molecular graph.
    \item It is directly adjacent to any atom satisfying either of the first two conditions.
    \item It belongs to the leaving group \(V_L\).
\end{itemize}
These extended criteria are adapted from retrosynthesis conventions, particularly following the methodology established in Retroformer~\cite{pmlr-v162-wan22a}.

\subsection{Atom-Aligned Encoder}
% Understanding chemical reaction mechanisms is fundamental to developing robust representations in cheminformatics. Despite significant advancements, our grasp of these mechanisms remains incomplete, and the annotation of reaction mechanisms necessitates considerable effort from chemical experts. To navigate this challenge, we have adopted a pragmatic approach. Currently, there are well-established tools \cite{schwaller2021extraction,chen2024precise} that can delineate the atomic correspondence between reactants and products in chemical reactions. Integrating this atomic mapping information into models can enhance the identification of similarities and differences between reactants and products, thereby improving the model's capacity to understand the evolution of chemical reactions. Driven by these considerations, we introduce the Atom-Aligned Encoder, a model engineered to assimilate both the chemical reaction and its associated atom-mapping data for effective encoding of chemical reactions.

The Atom-Aligned Encoder comprises a series of identical blocks that iteratively refine node and edge features, similar to the iterative process in Graph Neural Networks (GNNs). Each block contains two separate Message-Passing Neural Network (MPNN) layers operating on the reactant and product graphs, respectively. These layers incorporate node and edge features according to molecular structure to generate intermediate node features. Let \(H^{R(k)}\) and \(H^{P(k)}\) denote the reactant and product node features at layer \(k\), and \(E^{R(k)}\), \(E^{P(k)}\) denote their corresponding edge features. The intermediate node features are computed as follows:
\begin{equation}\label{eq: mpnn_in}
\begin{aligned}
    &\tilde{H}^{R(k)} =  {\rm MPNN}^{(k)}_1\left(H^{R(k-1)}, E^{R(k-1)}\right),\\
    &\tilde{H}^{P(k)} =  {\rm MPNN}^{(k)}_2\left(H^{P(k-1)}, E^{P(k-1)}\right).
\end{aligned}
\end{equation}
While standard graph neural networks process reactants and products independently, this decoupled architecture fails to capture the fine-grained structural changes between them; as a result, it cannot adequately model the underlying molecular transformations. To address this, we introduce an information fusion layer after each MPNN step that integrates the features of corresponding atom pairs. Specifically, the intermediate features of the leaving group atoms are processed through an auxiliary feedforward network:
\begin{equation}\label{eq: fusion}
\begin{aligned}
    & \left[h_i^{R(k)} \parallel h_i^{P(k)}\right] = {\rm FFN}_1\left(\left[\tilde{h}_i^{R(k)} \parallel \tilde{h}_i^{P(k)}\right]\right), &v^R_i, v_i^P \notin V_{L}, \\
    & h_i^{R(k)} = {\rm FFN}_2 \left(\tilde{h}_i^{R(k)}\right), &v^R_i \in V_L.\\
\end{aligned}
\end{equation}
This forced alignment between corresponding atoms enables the model to explicitly capture bond changes and atom conservation patterns, a key distinction from conventional GNN-based reaction encoders. Finally, edge features are updated based on the features of their endpoint nodes:
\begin{equation}\label{eq: eup}
\begin{aligned}
    & e_{ij}^{R(k)} = {\rm FFN}_3\left(\left[h_i^{R(k)} \parallel h_j^{R(k)}\right]\right), \\
    & e_{ij}^{P(k)} = {\rm FFN}_4\left(\left[h_i^{P(k)} \parallel h_j^{P(k)}\right]\right).
\end{aligned}
\end{equation}

\subsection{Adapter Module for Reaction Condition Integration} \label{sec: cond adapter}
Reaction conditions exhibit significant variability across datasets and may involve multiple modalities, such as reagent molecules or textual procedural descriptions~\cite{Wang2023,yoshikawa2023large}. Existing approaches typically concatenate condition features with molecular representations or employ fixed subnetwork architectures, which lack flexibility and cannot accommodate missing, non-molecular, or variably structured condition entries. To address these limitations, we design an adapter module inspired by T2IAdaptor~\cite{mou2024t2i} and ControlNet~\cite{zhang2023adding}, which injects condition information into the encoder via multi-head cross-attention without modifying the core graph architecture .

Each reaction condition entry  $C_i$ is first encoded into a feature matrix $\mathbf{C}_i \in \mathbb{R}^{c_i \times d}$ using a modality-specific encoder (e.g., a graph neural network for molecules, a multilayer perceptron for scalars, or a pretrained language model for text). The entry $C_i$ may represent a reagent molecule, a temperature scalar, or a text snippet. At every block $k$ of the Atom-Aligned Encoder, after the information fusion step and before the edge update, the node features are refined by cross-attention, whereby they serve as queries that attend to the condition matrices as keys and values: 
\begin{equation}\label{eq: cond}
    \begin{aligned}
        & \left[\dot{h}_i^{R(k)} \parallel \dot{h}_i^{P(k)}\right] = {\rm FFN}_1\left(\left[\tilde{h}_i^{R(k)} \parallel \tilde{h}_i^{P(k)}\right]\right), & v^R_i, v_i^P \notin V_{L} \\
        & \dot{h}_i^{R(k)} = {\rm FFN}_2 \left(\tilde{h}_i^{R(k)}\right), & v^R_i \in V_L \\
        & h_i^{R(k)} = \dot{h}_i^{R(k)} + \sum_{j=1}^c {\rm Attn}^{(1)}_j\left(\dot{h}_i^{R(k)}, \textbf{C}_j, \textbf{C}_j\right) \\
        & h_i^{P(k)} = \dot{h}_i^{P(k)} + \sum_{j=1}^c {\rm Attn}^{(2)}_j\left(\dot{h}_i^{P(k)}, \textbf{C}_j, \textbf{C}_j\right) \\
        % & e_{ij}^{R(k)} = {\rm FFN}_3\left(\left[h_i^{R(k)} \parallel h_j^{R(k)}\right]\right) \\
        % & e_{ij}^{P(k)} = {\rm FFN}_4\left(\left[h_i^{P(k)} \parallel h_j^{P(k)}\right]\right)
    \end{aligned}
\end{equation}
Here, $c$ denotes the number of condition entries, and ${\rm Attn}(Q,K,V)$ denotes standard multi-head attention~\cite{vaswani2017attention}:
\begin{equation}\label{eq: attn}
    \begin{aligned}
        & o_i = {\rm softmax}\left(\frac{QW_i^Q (KW_i^K)^\top}{\sqrt{d}}\right) VW_i^V \\
        & {\rm Attn}(Q, K, V) = \left[o_1 \Vert o_2 \Vert \cdots \Vert o_h\right] W^O,
    \end{aligned}
\end{equation}
where  $h$ is the number of heads, $d$ is the dimension of the key vectors, and $W_i^Q$, $W_i^K$, $W_i^V$, and $W^O$ are learnable parameters. The residual connection preserves the original node features, while the attention term dynamically incorporates context relevant to the condition. 

This design offers improved flexibility compared to prior reaction representation models that typically treat conditions as part of the reactants or rely on fixed subnetworks. It supports an arbitrary number and type of condition entries and can be readily combined with various pre-trained encoders, making it a practical choice for datasets with heterogeneous condition annotations.  Consequently, when employed as the backbone for large‑scale pretraining, our framework can naturally accommodate a broader spectrum of downstream tasks that involve previously unseen condition types, which is a critical advantage for building generalizable reaction foundation models.

\subsection{Reaction-Center-Aware Decoders}
The decoder utilizes encoded node features \(H^{R(L)}\) and \(H^{P(L)}\) from the Atom-Aligned Encoder's final layer (\(L\)-th blocks) to generate task-specific outputs. Central to our decoder design is a Reaction-Center-Aware (RC-aware) cross-attention mechanism that explicitly distinguishes reaction-center atoms from peripheral ones. This mechanism, adapted from the local-global decoder of Retroformer~\cite{pmlr-v162-wan22a}, employs two types of attention heads: half operate as standard global attention over all encoder nodes, while the other half attends exclusively to the reaction center set \(V_{rc}\). For an RC-constrained head, given a query vector $Q$, the attention distribution is computed only over the nodes in $V_{rc}$:
\begin{equation}\label{rcattn}
    \begin{aligned}
        &\alpha^1_l = \frac{\exp{(k_l^R q^\top)}}{\sum\limits_{v_j^R \in V_{rc}} \exp{(k_j^R q^\top)} + \sum\limits_{v_j^P \in V_{rc}} \exp{(k_j^P q^\top)}} \\
        &\alpha^2_l = \frac{\exp{(k_l^P q^\top)}}{\sum\limits_{v_j^R \in V_{rc}} \exp{(k_j^R q^\top)} + \sum\limits_{v_j^P \in V_{rc}} \exp{(k_j^P q^\top)}} \\
        & o_i = \sum_{v^R_l \in V_{rc}} \frac{\alpha_l^1}{\sqrt{d}} h^{R(L)}_l W^V_i + \sum_{v^P_l \in V_{rc}} \frac{\alpha_l^2}{\sqrt{d}} h^{P(L)}_l W^V_i \\
        & \left[k^R_l, k^P_l, q\right] = \left[h_l^{R(L)}W_i^K , h_l^{P(L)} W_i^K , QW_i^Q \right]
    \end{aligned}
\end{equation}
where \(W_i^Q\), \(W_i^K\), and \(W_i^V\) are learnable weights, and \(d\) is the key dimension. Global heads follow the same formulation but compute attention over all encoder nodes. The outputs of all heads are concatenated and linearly projected:
\begin{equation}
    {\rm RCAttn}(Q, H^{R(L)} \cup H^{P(L)}, V_{rc}) = \left[o_1 \Vert o_2 \Vert \cdots \Vert o_h\right] W^O
\end{equation}
with learnable weights \(W^O\) and \(h\) attention heads. Compared to the original Retroformer decoder, which considers only product atoms, we extend this formulation to include both reactants and products, thereby enabling the model to capture leaving-group interactions and bond-breaking events. Furthermore, we adapt this mechanism to two distinct task paradigms that encompass a wide range of reaction-related tasks.

For sequential generation tasks, where the target outputs are variable-length sequences of discrete tokens, such as in reaction condition recommendation, we replace the standard cross-attention mechanism in the Transformer decoder with our proposed reaction-center-aware (RC-aware) cross-attention. The decoder autoregressively generates the sequence of entry tokens, and the hidden state corresponding to each generated token serves as the query vector \(Q\) in Eq.~\ref{rcattn}. During the autoregressive decoding process, these query vectors are progressively concatenated to form an evolving query sequence.

For single-output tasks, such as yield prediction or other property prediction tasks, no explicit input query sequence is available. Instead, we introduce a learnable query vector \(Q_{\text{learn}}\in \mathbb{R}^d\), which is randomly initialized and optimized during training. This query vector is fed into the RC-aware cross-attention mechanism to produce a fixed-dimensional reaction representation. The resulting representation is then passed through a multilayer perceptron (MLP) to generate the final prediction.

\backmatter

% \bmhead{Acknowledgements}

% Acknowledgements are not compulsory. Where included they should be brief. Grant or contribution numbers may be acknowledged.

% Please refer to Journal-level guidance for any specific requirements.

\section*{Declarations}

\subsection*{Supplementary information}
Supplementary information is included after the main article in this arXiv version.

\subsection*{Data availability}
The raw data for USPTO-Condition can be found in \url{https://github.com/wangxr0526/Parrot}. The raw data for USPTO-500MT can be found in \url{https://yzhang.hpc.nyu.edu/T5Chem/}. The raw data for the Buchwald-Hartwig cross-coupling dataset can be found in \url{https://github.com/rxn4chemistry/rxn_yields}. The two selectivity datasets, the chiral C-H functionalization dataset and chiral phosphoric acid-catalyzed thiol addition dataset, can be found in \url{https://github.com/Masker-Li/ChemSelML} and \url{https://github.com/Shuwen-Li/SEMG-MIGNN/blob/main/Data/data2/data2.csv}, respectively. We also provide the raw data and our processed version in \url{https://github.com/zengkaipeng/AlignReact-Release}.

% The Data of USPTO-FULL and USPTO-50K can be found in \url{https://github.com/Hanjun-Dai/GLN}. The Data of USPTO-MIT can be found in \url{https://github.com/wengong-jin/nips17-rexgen/tree/master/USPTO}. We also provide the raw data and our processed version in \url{https://github.com/zengkaipeng/UAlign}.

\subsection*{Code availability}
All the code and checkpoints can be found in \url{https://github.com/zengkaipeng/AlignReact-Release}.

\subsection*{Author contribution}
K.Z. conceived the methodology, implemented the codebase, designed the algorithms, and authored the majority of the manuscript. X.L. reproduced the baseline methods, created a subset of figures, and contributed to the Results section. Y.Z. participated in model conceptualization, experimental design, and article polishing. Y.X. discussed the methodology and revised the manuscript. Y.X., Y.J., and X.Y. supervised the research.
\clearpage
\section*{Supplementary Information}
\addcontentsline{toc}{section}{Supplementary Information}
\setcounter{section}{0}
\setcounter{subsection}{0}
\setcounter{subsubsection}{0}
\setcounter{figure}{0}
\setcounter{table}{0}
\setcounter{equation}{0}
\renewcommand{\tablename}{Supplementary Table}
\renewcommand{\figurename}{Supplementary Fig.}
\renewcommand{\theHsection}{supp.\arabic{section}}
\renewcommand{\theHsubsection}{supp.\arabic{section}.\arabic{subsection}}
\renewcommand{\theHsubsubsection}{supp.\arabic{section}.\arabic{subsection}.\arabic{subsubsection}}
\renewcommand{\theHfigure}{supp.\arabic{figure}}
\renewcommand{\theHtable}{supp.\arabic{table}}
\renewcommand{\theHequation}{supp.\arabic{equation}}
\section{Data Preparation}\label{app: datapre}

\subsection{Descriptions and Statistical Information of Datasets}\label{app:data det}
The detailed descriptions of all datasets we used are presented as follows.

\begin{itemize} 
    \item \textbf{USPTO-Condition dataset}~\cite{Wang2023}. USPTO-Condition dataset is a dataset processed by Wang et al.~\cite{Wang2023}. This dataset comprises 680,741 reactions, with each reaction condition consistently consisting of up to one catalyst, two reagents, and two solvents.
    \item \textbf{USPTO-500MT dataset}~\cite{lu2022unified}. USPTO-500MT dataset is a dataset collected by Lu et al.~\cite{lu2022unified}. It can be used to train and test five different types of reaction tasks. In this paper, we only use it for the reaction condition combination generation tasks. 
    \item \textbf{Buchwald-Hartwig cross-coupling dataset}~\cite{ahneman2018predicting}. This dataset provides 10 random splits and four ligand-based out-of-sample splits for evaluation. The test sets under the four out-of-sample data splits contain reaction additives that are not included in the training sets. We use the raw data and the data split provided by Probst el al. ~\cite{probst2022reaction} and add the atom mapping for reactions according to the reaction template. 
    \item  \textbf{chiral C-H functionalization  dataset}~\cite{dataset1_hongxin_Angew_regio_li2020predicting}. The dataset contains 6,115 reactions that functionalize C-H bonds of heterocycles (via free radical reactions), with regioselectivity due to differences in reaction sites. The dataset were randomly divided into a training set and a test set with a 7: 3 ratio for 10 times.
    \item \textbf{chiral phosphoric acid-catalyzed thiol addition dataset}~\cite{dataset2_denmark_stereo_zahrt2019prediction,dataset2_hongxin_nc_stereo_li2023reaction}. The dataset contains the reactions regarding chiral phosphoric acid–catalyzed thiol addition to N-acylimines. It includes 43 catalysts and combinations of 5 $\times$ 5 reactants, which form 1075 reactions. The dataset were randomly divided into a training set and a test set with a 7: 3 ratio for 10 times.
\end{itemize}
We summarize the statistical information of all the datasets used in this work in Supplementary Table~\ref{tab: regression-dataset}.
\begin{table}[htbp]
    \centering
    
    \caption{\textbf{The statistical information of the datasets} }
    \label{tab: regression-dataset}
    \begin{tabular}{lcccc}
    \toprule
         \textbf{Dataset}  & Split-type  & Train & Val & Test \\
         \midrule
    USPTO-Condition & random split & 544,591&68,075&68,075\\
    USPTO-500MT & random split & 116,360 & 12,937 & 14,238\\
    \midrule
        \multirow{5}{*}{Buchwald-Hartwig cross-coupling} & random split & 2,491 & 277 & 1187 \\
         & Test1 & 2,751 & 306 & 898 \\
        & Test2 & 2,749 & 306 & 900 \\
        & Test3 & 2,752 & 306 & 897 \\
        & Test4 & 2,749 & 306 & 900 \\

         \midrule
         chiral phosphoric acid-catalyzed thiol addition & random split & 677 & 75 & 323 \\
         radical C-H functionalization & random split & 3,851 & 428 & 1,835 \\
       \bottomrule
    \end{tabular}
\end{table}

\subsection{Adding Atom Mapping}
The Buchwald-Hartwig cross-coupling dataset features a consistent reaction template across all its reaction data, enabling the derivation of atom mapping through rule-based approaches. For all the other datasets except the Buchwald-Hartwig cross-coupling dataset, we have employed RXNMapper~\cite{schwaller2021extraction} to obtain atom mapping. It is noteworthy that we have re-annotated the labels for the USPTO-500MT dataset, treating both the originally provided reactants and reagents as input reactants during the atom-mapping annotation for this dataset. For further details on the processing of the USPTO-500MT dataset, please refer to Supplementary Sec.~\ref{app: pre_500_mt}.

\subsection{Generation of Reagents in USPTO-500MT dataset}\label{app: pre_500_mt}
We restructured the USPTO-500MT dataset. Following the annotations provided by the dataset, we divided the reactants involved in the reactions into a series of charge-balanced ions or molecular clusters. Here, we treat a cluster as a single molecule rather than segmenting molecules based on the delimiter `.' as in prior studies. Utilizing RXNMapper~\cite{schwaller2021extraction}, we appended atom mapping to the reactions, where molecules in the original reactants that have atoms present in the products are considered reactants, while the remaining portions are classified as reagents. We categorized the reagents according to the following hierarchy:
\begin{itemize}
\item If a molecule is a free metal, contains a cyclic structure along with a metal or phosphorus atom, or is a metal halide, it is designated as Type I.
\item If a molecule is an organic compound, it is designated as Type II.
\item The remaining molecules are categorized as Type III.
\end{itemize}
Our data labels are constructed in the order of Type I, Type II, and Type III. When reagents contain multiple molecules of the same type, these molecules are sorted in ascending order of their SMILES string lengths within the same category. The tokenization of the SMILES representations of the reagents is aligned with the work~\cite{schwaller2019molecular}.

\section{The Full Experiment Results}\label{sec:si_full_results}
\subsection{Reaction Condition Recommendation}
Experimental results for the USPTO-Condition and USPTO-500MT datasets are tabulated in Supplementary Tables~\ref{tab: res_uspto_condition}, Supplementary Table~\ref{tab:res_uspto_condition_seperate}, and Supplementary Table~\ref{tab: res_uspto_500MT}. For USPTO-Condition, we additionally report the prediction accuracies for catalysts, solvents, and reagents. Each entry may contain up to two solvents and two reagents; crucially, the prediction order is not considered in calculating the solvents'and reagent's prediction accuracy.

\begin{table}[htbp]
\centering
% \begin{minipage}{0.49\textwidth}
    \centering
    
    \caption{\textbf{Overall accuracy of reaction condition prediction on USPTO-Condition dataset.} The best performance is in \textbf{bold}. The second-best performance is underlined. * indicates that the model has been pretrained.  IFL stands for information fusion layer.}
    \label{tab: res_uspto_condition}
    % \resizebox{\linewidth}{!}{
    \begin{tabular}{lcccc}
    \toprule
       \multicolumn{1}{c}{\multirow{2}{*}{\textbf{Model}}}  &  \multicolumn{4}{c}{\textbf{Top-$k$ accuracy (\%)}} \\
       \cmidrule{2-5}
          & 1&3&5&10\\
          \midrule
          % ~\cite{maser2021multilabel}, ~\cite{Gao2018}, ~\cite{andronov2023reagent}, ~\cite{Wang2023}
          T5chem$^*$ & \underline{28.31} & \underline{42.32} &  \underline{48.15} & \underline{55.32} \\
          Parrot$^*$ %~\cite{Wang2023}
            & 27.42 & 41.86 & 46.98 & 50.95\\
        \midrule
          GCNN %~\cite{maser2021multilabel}
            & 12.81 & 21.95 & 26.40 & 32.17 \\
          FPRCR %~\cite{Gao2018}
            & 16.90 & 26.38 & 31.16 & 36.96\\
          Reagent  Transformer%~\cite{andronov2023reagent}
            & 22.74 & 34.09 & 39.36 & 46.01 \\
          % Parrot~\cite{Wang2023} & 26.91 & 40.35 & 45.10 & 49.14\\
            T5Chem & 16.42	&27.75	&33.53&	40.79\\
            
     Parrot\tnote{1} &23.97 & 36.72& 41.59 & 45.96\\
          Ours&  \textbf{33.30}& \textbf{47.11} &  \textbf{52.43} & \textbf{59.19}\\
          % Ours & \textbf{34.30}&	\textbf{47.60}&	\textbf{52.82} & 	\textbf{59.22}\\
          % Ours - IFL & 32.76 & 46.49 & 52.06& 58.79\\
        \bottomrule
    \end{tabular}
    \begin{tablenotes}
    \item[1] The checkpoint of non-pretrained version is not provided, the performance is copied from the original paper~\cite{Wang2023}.
    \end{tablenotes}
    % }
\end{table}

\begin{sidewaystable}[htbp]
\centering

\caption{\textbf{Accuracy of each component on USPTO-Condition dataset for reaction condition prediction.} The detailed prediction accuracy on each type of components is displayed. The best performance is in \textbf{bold} and the second-best is underlined. *indicates the the model is pretrained}
    \begin{tabular}{lcccccccccccccc}
    \toprule
    \multicolumn{1}{c}{\multirow{3}*{Method}} & \multicolumn{14}{c}{\textbf{Top-$k$ accuracy (\%)}}\\
    \cmidrule{2-15}
    & \multicolumn{4}{c}{catalyst} & & \multicolumn{4}{c}{solvents} & & \multicolumn{4}{c}{reagents} \\
    \cmidrule{2-5} \cmidrule{7-10} \cmidrule{12-15}   & 1 & 3 & 5 & 10 & & 1 & 3 & 5 & 10 & & 1 & 3 & 5 & 10 \\
    \midrule
    % parrot：~\cite{Wang2023}， GCNN~\cite{maser2021multilabel}， Reag. Trans, fprcr ~\cite{Gao2018}, 
    {Parrot$^*$} % ~\cite{Wang2023}
        & \underline{92.12} & \underline{94.91} & \underline{95.97} & \underline{97.28} & & \underline{44.20} & \underline{59.23} & \underline{63.71} & \underline{66.16} & & \underline{46.47} & \underline{62.04} & \underline{68.19} & \underline{73.93} \\
    {T5Chem$^*$} % [pretrain]
     & 91.64 & 94.76 & 95.88 & 97.07 & & 44.17 & 59.02 & 65.14 & 72.95 & & 46.23 & 62.37 & 68.55 & 75.66 \\
     \midrule
   {Reagent Transformer} %~\cite{andronov2023reagent}
        & 89.80 & 93.30 & 94.52 & 95.88 & & 37.97 & 51.29 & 57.47 & 65.44 & & 39.11 & 55.20 & 61.83 & 69.80 \\
   {FPRCR} % ~\cite{Gao2018}
        & 91.22 & 92.82 & 93.57 & 94.74 & & 38.51 & 49.72 & 54.64 & 61.56 & & 37.54 & 47.95 & 52.69 & 58.48 \\
    {GCNN} % ~\cite{maser2021multilabel}
        & 90.59 & 91.80 & 92.40 & 93.39 & & 32.39 & 46.02 & 52.51 & 61.06 & & 35.84 & 46.37 & 50.61 & 55.99 \\
    
    % [from scratch]
     T5Chem & 88.60 & 92.38 & 93.68 & 95.07 & & 30.30 & 46.66 & 54.63 & 64.57 & & 32.71 & 47.98 & 54.86 & 63.25 \\
     Parrot\tnote{1} & 91.96 & 91.96 & 91.96 & 91.96 && -&-&-&-&& -&-&-&-\\
   Ours & \textbf{92.65} & \textbf{95.56}& \textbf{96.46} & \textbf{97.47} && \textbf{48.60} & \textbf{63.09} & \textbf{68.81} & \textbf{75.76} && \textbf{49.68} & \textbf{66.34} & \textbf{72.28} & \textbf{79.02}\\
    \bottomrule
    \end{tabular}
\begin{tablenotes}
\item[1] The checkpoint of non-pretrained version is not provided, the prediction accuracy on solvents and reagents can not be calculated. The prediction accuracy on catalyst is copied from the original paper~\cite{Wang2023}.
\end{tablenotes}
\label{tab:res_uspto_condition_seperate}
\end{sidewaystable}

\begin{table}[htbp]
    \centering
    
    \caption{\textbf{Overall accuracy of reaction condition generation on USPTO-500MT dataset.} The best performance is in \textbf{bold}. The second-best performance is underlined. * indicates that the
model has been pretrained.}
    \label{tab: res_uspto_500MT}
    % \resizebox{\linewidth}{!}{
   \begin{tabular}{lcccc}
    \toprule
       \multicolumn{1}{c}{\multirow{2}{*}{\textbf{Model}}}  &  \multicolumn{4}{c}{\textbf{Top-$k$ accuracy (\%)}} \\
       \cmidrule{2-5}
          & 1&3&5&10\\
          \midrule
          % t5chem: ~\cite{lu2022unified}, GCNN~\cite{maser2021multilabel},Reagent  Transformer~\cite{andronov2023reagent}, 
          T5Chem$^*$ %~\cite{lu2022unified}
            & \underline{25.36} & \underline{38.12} & \underline{43.42} & \underline{49.54} \\
          \midrule
          % GCNN %~\cite{maser2021multilabel}
          %   & 4.74&	12.36& 17.94&	24.58\\
          % GCNN*~\cite{maser2021multilabel} & \underline{20.86}&	\underline{35.68}&	\underline{42.72}&	\textbf{52.05} \\
          Reagent  Transformer %~\cite{andronov2023reagent}
            & {17.88}&	{25.47}&	{28.78}&	{33.27}\\
          % Ours & \textbf{26.08}&	\textbf{39.16}&	\textbf{44.38}&	\underline{50.78}\\
          T5Chem %~\cite{lu2022unified}
            & {17.32} & {28.21} & {32.98} & {38.76} \\
          Ours & \textbf{26.62}&	\textbf{39.46}&	\textbf{44.69 }&	\textbf{50.94}\\
        \bottomrule
    \end{tabular}
\end{table}

\subsection{Reaction Yield Prediction}\label{sec:si_yield_results}
We evaluated our model across ten random data splits and four out-of-distribution (OOD) data splits on the Buchwald-Hartwig cross-coupling dataset. We use MAE, RMSE, and $R^2$ as evaluation metrics. The results are summarized in Supplementary Table~\ref{tab: bh-iid},  Supplementary Table~\ref{tab: bh-ood1}, and Supplementary Table~\ref{tab: bh-ood2}.  For the random data splits, we report the mean and standard deviation across the ten splits. For the four OOD data splits, we conduct experiments using five random seeds (2023 to 2027) and present the mean and standard deviation of the results.

\begin{table}[htbp]
    \centering
    
    \caption{\textbf{The results on Buchwald-Hartwig cross-coupling dataset under ten random splits.} The best performance of pretrained and not pretrained model is in \textbf{bold} and the second-best performance of pretrained and not pretrained models is underlined. * indicates that the model has been pretrained, and $\dag$ indicates that our model uses a pretrained condition encoder. }\label{tab: bh-iid}
    \begin{tabular}{l c ccc}
    \toprule
          \textbf{Model} & \textbf{Type}  & MAE $\downarrow$ & RMSE $\downarrow$ & $R^2 \uparrow$ \\
         \midrule
        MFF %~\cite{MFF_SANDFORT20201379}
            &ML & 6.4411  $\pm$  0.8157 & 9.6382  $\pm$  1.0206 & 0.8739  $\pm$  0.0292 \\
          DRFP %~\cite{probst2022reaction}
             &ML& 4.0995  $\pm$  0.1191& 6.2424  $\pm$  0.2636& 0.9474  $\pm$  0.0050\\
          Chemprop %~\cite{chemprop:doi:10.1021/acs.jcim.3c01250}
            &DL & 4.6430  $\pm$  0.1405 & 6.4306  $\pm$  0.1938 & 0.9441  $\pm$  0.0042 \\
        YieldBert  %~\cite{schwaller2021prediction}
            % & {3.5532  $\pm$  0.1600} & {5.4480  $\pm$  0.3240}& {0.9599  $\pm$  0.0050}\\    
            	&DL& 10.9327  $\pm$  0.6514 & 15.7460  $\pm$  0.6475 & 0.6655  $\pm$  0.0311 \\
        T5Chem %~\cite{lu2022unified} 
            % & \textbf{3.5059  $\pm$  0.1562} & \textbf{5.3181  $\pm$  0.2482}& \textbf{0.9662  $\pm$  0.0034} \\
           &DL &	11.2666  $\pm$  0.6432 &  15.8840  $\pm$  0.7096 & 0.6601 $\pm$ 0.0253 \\
             ReaMVP &DL& \textbf{3.3921 $\pm$ 0.1453}	&	\textbf{5.1034 $\pm$ 0.2717} & \textbf{0.9648 $\pm$ 0.0039} \\
            Ours&DL & \underline{3.3972 $\pm$ 0.1530} & \underline{{5.1582 $\pm$ 0.2162}} & \underline{0.9642 $\pm$ 0.0026} \\
        \midrule
         YieldBert$^*$   %~\cite{schwaller2021prediction}
            & DL& {3.5532  $\pm$  0.1600} & {5.4480  $\pm$  0.3240}& {0.9599  $\pm$  0.0050}\\
        T5Chem$^*$  %~\cite{lu2022unified} 
            & DL& \underline{3.5059  $\pm$  0.1562} & \underline{5.3181  $\pm$  0.2482}& \underline{0.9607 $\pm$ 0.0047} \\
             ReaMVP$^*$ &DL& 3.7213  $\pm$   0.0413 & 5.6183   $\pm$   0.2772  & 0.9574  $\pm$  0.1637\\
         Ours$^\dag$ &DL & \textbf{3.2763 $\pm$ 0.1313} & \textbf{5.0662 $\pm$ 0.1609} &\textbf{ 0.9654 $\pm$ 0.0021} \\
           \bottomrule
    \end{tabular}
\end{table}

\begin{table*}[htbp]
    \centering
    
    \caption{\textbf{The results on Buchwald-Hartwig cross-coupling dataset under out-of-distribution data split Test1 and Test2.}  The best performance of pretrained and not pretrained model is in \textbf{bold} and the second-best performance of pretrained and not pretrained models is underlined. * indicates that the model has been pretrained, and $\dag$ indicates that our model uses a pretrained condition encoder. }\label{tab: bh-ood1}
    \resizebox{\linewidth}{!}{
    \begin{tabular}{lc  cccc cccc cccc cccc}
    \toprule
         \multicolumn{1}{c}{\multirow{2}{*}{\textbf{Model}}} & \multicolumn{1}{c}{\multirow{2}{*}{\textbf{Type}}}  & \multicolumn{3}{c}{\textbf{Test1}}&  &\multicolumn{3}{c}{\textbf{Test2}} \\
         \cmidrule{3-5}  \cmidrule{7-9} 
         &  &MAE $\downarrow$ & RMSE $\downarrow$ & $R^2 \uparrow$& &MAE $\downarrow$ & RMSE $\downarrow$ & $R^2 \uparrow$ \\
         \midrule
         % MFF~\cite{MFF_SANDFORT20201379}, DRFP ~\cite{probst2022reaction} , chemprop ~\cite{chemprop:doi:10.1021/acs.jcim.3c01250}, yieldBert ~\cite{schwaller2021prediction}, t5hcem ~\cite{lu2022unified},
         MFF %~\cite{MFF_SANDFORT20201379}
            & ML & 10.9730 $\pm$ 1.1066 & 17.4750 $\pm$ 2.9942 & 0.5795 $\pm$ 0.1364 && 9.6391 $\pm$ 0.1465  & 14.0512 $\pm$ 0.2180 & 0.7309 $\pm$ 0.0083 \\
         DRFP %~\cite{probst2022reaction}
            & ML &  \textbf{8.0137 $\pm$ 0.0702} & \textbf{11.3992 $\pm$ 0.0845}	& \textbf{0.8252 $\pm$ 0.0026} &&  \underline{9.0642 $\pm$ 0.0725}  & \underline{13.5332 $\pm$ 0.1913} & \underline{0.7504 $\pm$ 0.0070} \\
        
         Chemprop %~\cite{chemprop:doi:10.1021/acs.jcim.3c01250}
            & DL & 9.7982 $\pm$ 0.9586 & \underline{14.0620 $\pm$ 1.3639} & \underline{0.7320 $\pm$ 0.0523} &&10.0890 $\pm$ 0.7180 & 13.6659 $\pm$ 0.8532 & 0.7447 $\pm$ 0.0308 \\
        YieldBert & DL %~\cite{schwaller2021prediction}
              &  13.2461 $\pm$ 0.2341&18.8416 $\pm$ 0.4287&0.5222 $\pm$ 0.0216 && 13.0776 $\pm$ 0.9276&18.9203 $\pm$ 1.4600&0.5101 $\pm$ 0.0751 \\
           % T5Chem[pretrain]~\cite{lu2022unified} &  7.5644 & 12.1551 & 0.8128 &&  6.1671 & 9.4033 & 0.8919 &&  8.6164 & 13.2639 & 0.8053 &&  13.961 & 20.4851 & 0.5614 \\
          T5Chem%[multiTask]
           %~\cite{lu2022unified} 
           & DL & 12.3271 $\pm$ 0.5404&17.1757 $\pm$ 0.7090&0.6263 $\pm$ 0.0302 && 12.7461 $\pm$ 0.6414&17.8660 $\pm$ 0.8463&0.5896 $\pm$ 0.0333 \\
            % &  \underline{7.1283} & 11.2242 & \underline{0.8385} &&  \underline{6.7693} & \underline{10.4041} & \underline{0.8801} &&  \underline{9.0982} & \underline{14.3431} & \underline{0.7665} &&  13.4069 & 19.7521 & \textbf{0.6051} \\

         ReaMVP & DL & 13.2885 $\pm$ 1.3030&18.6288 $\pm$ 2.0344&0.5275 $\pm$ 0.0963 && 9.9564 $\pm$ 0.7027&15.5496 $\pm$ 0.6603&0.6700 $\pm$ 0.0281\\
           Ours& DL & \underline{8.5583 $\pm$ 1.6501} & 14.3919 $\pm$ 2.7852 & 0.7130 $\pm$ 0.1183 && \textbf{8.0124 $\pm$ 1.9118} & \textbf{12.2923 $\pm$ 3.2324} &\textbf{ 0.7827 $\pm$ 0.1025} \\
           \midrule
        
         YieldBert$^*$ %~\cite{schwaller2021prediction}
            &DL&  7.7793 $\pm$ 0.4532 & \underline{11.5510 $\pm$ 0.6199} & 0.8201 $\pm$ 0.0197 && 7.0805 $\pm$ 0.8213 & 10.4167 $\pm$ 1.3162 & 0.8503 $\pm$ 0.0371\\
           % T5Chem[pretrain]~\cite{lu2022unified} &  7.5644 & 12.1551 & 0.8128 &&  6.1671 & 9.4033 & 0.8919 &&  8.6164 & 13.2639 & 0.8053 &&  13.961 & 20.4851 & 0.5614 \\
            T5Chem$^*$%[multiTask] 
           %~\cite{lu2022unified}
            &DL& \underline{7.3660 $\pm$ 0.2220} & 12.0129 $\pm$ 0.2372 & \underline{0.8058 $\pm$  0.0077} 
 && \underline{6.1430 $\pm$ 0.2137} & \underline{9.4270 $\pm$ 0.3318} & \underline{0.8788 $\pm$ 0.0086} \\
         ReaMVP$^*$ & DL & 9.7030 $\pm$ 0.9769&14.3139 $\pm$ 1.1189&0.7227 $\pm$ 0.0449 &&  7.9062 $\pm$ 0.2206&12.2573 $\pm$ 0.6379&0.7947 $\pm$ 0.0217 \\
           % Ours$^\dag$& DL  &
           % \textbf{6.6177$\pm$1.0048} & \textbf{10.2934$\pm$1.5926} & \textbf{0.8547$\pm$0.0445}
           % && \textbf{5.5398$\pm$0.2160} & \textbf{8.3805$\pm$0.2856}& \textbf{0.9042$\pm$0.0065} \\
           Ours$^\dag$ & DL & \textbf{6.6637 $\pm$ 0.3786}	& \textbf{10.8275 $\pm$ 0.8035}&	\textbf{0.8414 $\pm$ 0.0236}	
 && \textbf{5.5398 $\pm$ 0.1932} &	\textbf{8.3805 $\pm$ 0.2555}&	\textbf{0.9042 $\pm$ 0.0059} \\
           \bottomrule
    \end{tabular}
    }
\end{table*}

\begin{table*}[htbp]
    \centering
    
    \caption{\textbf{The results on Buchwald-Hartwig cross-coupling  dataset under out-of-distribution data split Test3 and Test4.}  The best performance of pretrained and not pretrained model is in \textbf{bold} and the second-best performance of pretrained and not pretrained models is underlined. * indicates that the model has been pretrained, and $\dag$ indicates that our model uses a pretrained condition encoder. }\label{tab: bh-ood2}
    \resizebox{\linewidth}{!}{
    \begin{tabular}{lc  cccc cccc cccc cccc}
    \toprule
         \multicolumn{1}{c}{\multirow{2}{*}{\textbf{Model}}} & \multicolumn{1}{c}{\multirow{2}{*}{\textbf{Type}}}  & \multicolumn{3}{c}{\textbf{Test3}}&  &\multicolumn{3}{c}{\textbf{Test4}} \\
         \cmidrule{3-5}  \cmidrule{7-9} 
         &  &MAE $\downarrow$ & RMSE $\downarrow$ & $R^2 \uparrow$& &MAE $\downarrow$ & RMSE $\downarrow$ & $R^2 \uparrow$ \\
         \midrule
         % MFF~\cite{MFF_SANDFORT20201379}, DRFP ~\cite{probst2022reaction} , chemprop ~\cite{chemprop:doi:10.1021/acs.jcim.3c01250}, yieldBert ~\cite{schwaller2021prediction}, t5hcem ~\cite{lu2022unified},
         MFF %~\cite{MFF_SANDFORT20201379}
            & ML &\textbf{9.6022 $\pm$ 0.0996}&\textbf{14.8230 $\pm$ 0.2124}&\textbf{0.7216 $\pm$ 0.0080}&&15.9437 $\pm$ 2.0114&20.9527 $\pm$ 1.7867&0.3688 $\pm$ 0.1086 \\
         DRFP %~\cite{probst2022reaction}
            & ML &  \underline{10.1413 $\pm$ 0.0914} & {15.9365 $\pm$ 0.0954} &0.6782 $\pm$ 0.0039&&\textbf{12.7281 $\pm$ 0.0332}&\textbf{19.0366 $\pm$ 0.0644}&\textbf{0.4819 $\pm$ 0.0035} \\
        
         Chemprop %~\cite{chemprop:doi:10.1021/acs.jcim.3c01250}
            & DL & 10.2583 $\pm$ 0.4631& \underline{15.2133 $\pm$ 0.5174}&\underline{0.7065 $\pm$ 0.0197}&& \underline{14.5552 $\pm$ 0.7769}& \underline{20.3402 $\pm$ 0.9106} & \underline{0.4076 $\pm$ 0.0522} \\
        YieldBert & DL &  %~\cite{schwaller2021prediction}
        15.4959 $\pm$ 0.5699&22.1421 $\pm$ 0.8237&0.3783 $\pm$ 0.0462 && 17.7884 $\pm$ 0.9201&23.2340 $\pm$ 0.9895&0.2273 $\pm$ 0.0656
               \\
           % T5Chem[pretrain]~\cite{lu2022unified} &  7.5644 & 12.1551 & 0.8128 &&  6.1671 & 9.4033 & 0.8919 &&  8.6164 & 13.2639 & 0.8053 &&  13.961 & 20.4851 & 0.5614 \\
          T5Chem%[multiTask]
           %~\cite{lu2022unified} 
           & DL & 11.3490 $\pm$ 0.7840&16.7147 $\pm$ 0.4914&0.6566 $\pm$ 0.0185 && 20.6032 $\pm$ 0.7013&26.8503 $\pm$ 1.0382&0.2284 $\pm$ 0.0426 \\
            % &  \underline{7.1283} & 11.2242 & \underline{0.8385} &&  \underline{6.7693} & \underline{10.4041} & \underline{0.8801} &&  \underline{9.0982} & \underline{14.3431} & \underline{0.7665} &&  13.4069 & 19.7521 & \textbf{0.6051} \\

         ReaMVP & DL & 13.0350 $\pm$ 1.8263 & 19.7192 $\pm$ 1.9116 &0.5028 $\pm$ 0.0962 && 18.2115 $\pm$ 2.6947 &24.7563 $\pm$ 3.4819 &0.1065 $\pm$ 0.2614 \\
           % Ours & DL & 10.6888 $\pm$ 0.4226 & 17.7071 $\pm$	0.7213 & 0.6021 $\pm$	0.0327 && 15.0192 $\pm$	0.6314 & 21.4283 $\pm$	0.9210 & 0.3424 $\pm$ 0.0564 \\
           Ours & DL & 10.8734 $\pm$ 0.4776 & 17.7394 $\pm$ 0.8088 & 0.6007 $\pm$ 0.0367 && 15.7305 $\pm$ 0.5098 & 23.5543 $\pm$ 0.7971 & 0.2062 $\pm$ 0.0535 \\
           \midrule
        
         YieldBert$^*$ %~\cite{schwaller2021prediction}
            &DL&  9.6526 $\pm$ 0.1736&15.4455 $\pm$ 0.7289&0.6972 $\pm$ 0.0286&&14.3879 $\pm$ 1.1374&19.6979 $\pm$ 1.5110&0.4427 $\pm$ 0.0834\\
           % T5Chem[pretrain]~\cite{lu2022unified} &  7.5644 & 12.1551 & 0.8128 &&  6.1671 & 9.4033 & 0.8919 &&  8.6164 & 13.2639 & 0.8053 &&  13.961 & 20.4851 & 0.5614 \\
            T5Chem$^*$%[multiTask]
           %~\cite{lu2022unified}
            &DL& \underline{9.1328 $\pm$ 0.4950}& 14.4337 $\pm$ 1.3158&\underline{0.7339 $\pm$ 0.0485}&& \underline{13.5883 $\pm$ 0.5530}& 19.7144 $\pm$ 0.6983 & \underline{0.4437 $\pm$ 0.0393}\\
         ReaMVP$^*$ & DL & 9.7782  $\pm$  0.2469&\underline{14.5569 $\pm$ 0.2282}&0.7315 $\pm$ 0.0084  && \textbf{12.6095 $\pm$ 1.0512} & \textbf{17.8684 $\pm$ 0.7946}&\textbf{0.5420 $\pm$ 0.0531}\\
          % old
           % Ours$^\dag$& DL & \textbf{8.6564 $\pm$	0.4237} & \textbf{12.7284 $\pm$	0.6701} & \textbf{0.7942 $\pm$ 0.0214 }&& \textbf{12.3490 $\pm$	0.4484} & \underline{18.2576 $\pm$	0.6810} & {0.5228 $\pm$	0.0355}   \\
           Ours$^\dag$& DL & \textbf{8.9069 $\pm$ 0.4436} & \textbf{13.8243 $\pm$ 0.8078} & \textbf{0.7572 $\pm$ 0.0282} && 
           14.0090 $\pm$ 0.2941 &	20.3306 $\pm$ 0.5318&	0.4087 $\pm$ 0.0311 \\
           \bottomrule
    \end{tabular}
    }
\end{table*}

\subsection{Reaction Selectivity Prediction}\label{sec:si_selectivity_results}
We evaluated our model's performance on both the thiol addition and C–H functionalization datasets. Each dataset was partitioned into ten distinct random splits. Using MAE, RMSE, and $R^2$ as evaluation metrics, we report the mean and standard deviation of these metrics across all ten splits for each dataset. These results are summarized in Supplementary Table~\ref{tab: selectivity_1} and Supplementary Table~\ref{tab: selectivity_2}. We also report the performance of different models on three out-of-distribution test sets of the thiol addition dataset. Five random seeds (2025 to 2029) were used for training, and the mean and standard deviation of all metrics are reported in Supplementary Table~\ref{tab: sel2_cat}, Supplementary Table~\ref{tab: sel2_sub}, and Supplementary Table~\ref{tab: sel2_cat_sub}.

\begin{table}[htbp]
    \centering
    
    \caption{\textbf{The results on radical C-H functionalization dataset under random splits.} The best performance is in \textbf{bold}. The second-best performance is underlined. * indicates that the model has been pretrained.}
    \label{tab: selectivity_1}
    % \resizebox{\linewidth}{!}{
    \begin{tabular}{l cccc}
    \toprule
         \textbf{Model} & \textbf{Type} & MAE $\downarrow$ & RMSE $\downarrow$ & $R^2 \uparrow$ \\
         \midrule
         RXNFP$^*$ %~\cite{schwaller2021mapping} 
            &DL & \textbf{0.3744 $\pm$ 0.0094} & \textbf{0.5378 $\pm$ 0.0204} & \textbf{0.9862 $\pm$ 0.0010} \\
         T5Chem$^*$ %~\cite{lu2022unified} 
            &DL& \underline{0.5930 $\pm$ 0.0148} & \underline{0.7912 $\pm$ 0.0219}  & \underline{0.9702 $\pm$ 0.0016}  \\
        ReaMVP$^*$&DL& 0.5908 $\pm$ 0.0099 & 0.8237 $\pm$ 0.015 & 0.9677 $\pm$ 0.0012 \\
        \midrule
         % Chemma &   & 1.33  & 0.93\\
         MFF %~\cite{MFF_SANDFORT20201379} 
            &ML& 2.2923 $\pm$ 0.0400 & 2.9388 $\pm$ 0.0393 & 0.5891 $\pm$ 0.0095 \\
         DRFP %~\cite{probst2022reaction}
            &ML& 0.9435 $\pm$ 0.0293 & 1.2943 $\pm$ 0.0387 & 0.9203 $\pm$ 0.0042 \\
         % MIGNN~\cite{dataset2_hongxin_nc_stereo_li2023reaction} & 1.6886 $\pm$ 0.2199 & 2.2265 $\pm$ 0.2745 & 0.7610 $\pm$ 0.0591 \\
         % GAT_MIGNN~\cite{dataset2_hongxin_nc_stereo_li2023reaction} & 0.7251 $\pm$ 0.0822 & 0.9765 $\pm$ 0.1135 & 0.9542 $\pm$ 0.0109 \\
         Chemprop % ~\cite{chemprop:doi:10.1021/acs.jcim.3c01250} 
            &DL& \underline{0.3593 $\pm$ 0.0099} & \underline{0.5396 $\pm$ 0.0299} & \underline{0.9861 $\pm$ 0.0015}\\
         RXNFP %~\cite{schwaller2021mapping} 
            &DL& 1.0034 $\pm$ 0.0393 & 1.4389 $\pm$ 0.0638 & 0.9013 $\pm$ 0.0089   \\
         T5Chem %~\cite{lu2022unified} 
              &DL & 2.5398 $\pm$ 0.0549 & 3.3156 $\pm$ 0.0925 & 0.4765 $\pm$ 0.0322 \\
         % Ours & \textbf{0.3267 $\pm$ 0.0159} & \textbf{0.5110 $\pm$ 0.0292} & \textbf{0.9875 $\pm$ 0.0014} \\ % dim:128-heads:8-layer:5-dp:0.0-bs:64 best ep < 300(/400)
         % Ours  &  0.4158 $\pm$ 0.02142 & 0.5885 $\pm$ 0.0281 & 0.9835 $\pm$ 0.0015 \\ % 128	8	3, lr=0.0005 dp=0.1; best ep < 312(/400)
         ReaMVP&DL& 0.6390 $\pm$ 0.0216 & 0.8653 $\pm$ 0.0307 & 0.9643 $\pm$ 0.0026\\
         Ours &DL& \textbf{0.3362 $\pm$ 0.0158} & \textbf{0.5375 $\pm$ 0.0379} & \textbf{0.9862 $\pm$ 0.0018}\\
           \bottomrule
    \end{tabular}
    % }
\end{table}

\begin{table}[htbp]
    \centering
    
    \caption{\textbf{The results on chiral phosphoric acid-catalyzed thiol addition dataset under random splits.} The best performance is in \textbf{bold}. The second-best performance is underlined.}
    \label{tab: selectivity_2}
    % \resizebox{\linewidth}{!}{
    \begin{tabular}{l cccc}
    \toprule
         \textbf{Model} & \textbf{Type} & MAE $\downarrow$ & RMSE $\downarrow$ & $R^2 \uparrow$ \\
         \midrule
           
         % Chemma &   & 0.25  & 0.89\\
         MFF %~\cite{MFF_SANDFORT20201379} 
          & ML & \textbf{0.1421 $\pm$ 0.0093} & \underline{0.2122 $\pm$ 0.0169} & \textbf{0.9055 $\pm$ 0.0119} \\
         DRFP %~\cite{probst2022reaction} 
           &ML & \underline{0.1460 $\pm$ 0.0086} & \textbf{0.2109 $\pm$ 0.0141} & \underline{0.9066 $\pm$ 0.0115}\\
         % MIGNN~\cite{dataset2_hongxin_nc_stereo_li2023reaction} & 0.1731 $\pm$ 0.0106 & 0.2602 $\pm$ 0.0286 & 0.8578 $\pm$ 0.0251 \\
         Chemprop % ~\cite{chemprop:doi:10.1021/acs.jcim.3c01250} 
           &DL & 0.1626 $\pm$ 0.0118 & 0.2283 $\pm$ 0.0174 & 0.8907 $\pm$ 0.0130 \\
          RXNFP % ~\cite{schwaller2021mapping} 
            &DL& 0.2717 $\pm$ 0.0150 & 0.3935 $\pm$ 0.0223 & 0.6747 $\pm$ 0.0368  \\
            T5Chem &DL&  0.3941 $\pm$ 0.0287 & 0.5177 $\pm$ 0.0459 & 0.4293 $\pm$ 0.1344 \\ %~\cite{lu2022unified} \\
           % Ours\_old & 0.1535 $\pm$ 0.0085 & 0.2202 $\pm$ 0.0118  & 0.8982 $\pm$ 0.0103\\ % 128	8 3, lr=5.00E-05; dp =0;  pregnn 300	5; best ep: <200 (/400) 
           ReaMVP&DL& 0.1920 $\pm$ 0.0889 & 0.2721 $\pm$ 0.1219 & 0.8127 $\pm$ 0.2325  \\
           Ours &DL& 0.1580 $\pm$ 0.0079  &  0.2249 $\pm$ 0.0117  & 0.8937 $\pm$ 0.0115 \\
           \midrule
           RXNFP$^*$ % ~\cite{schwaller2021mapping} 
            &DL& {0.1650 $\pm$ 0.0111} & \underline{0.2313 $\pm$ 0.0197} & \underline{0.8872 $\pm$ 0.0193} \\
          T5Chem$^*$ %~\cite{lu2022unified} 
            &DL& 0.1812 $\pm$ 0.0128 & 0.2614 $\pm$ 0.0213 & 0.8568 $\pm$ 0.0171\\
           ReaMVP$^*$&DL& \underline{0.1647 $\pm$ 0.0073} & {0.2373 $\pm$ 0.0146} &  {0.8817 $\pm$ 0.0142} \\
           Ours$^\dag$ & DL& \textbf{0.1531 $\pm$ 0.0094} & \textbf{0.2215 $\pm$ 0.0147 }& \textbf{0.8970 $\pm$ 0.0117}\\
           \bottomrule
    \end{tabular}
    % }
\end{table}

\begin{table}[htbp]
    \centering
    
    \caption{\textbf{The results on chiral phosphoric acid-catalyzed thiol addition dataset(Cat test set).} The best performance of pretrained and not pretrained model is in \textbf{bold} and the second-best performance of pretrained and not pretrained models is underlined. * indicates that the model has been pretrained, and $\dag$ indicates that our model uses a pretrained condition encoder. }\label{tab: sel2_cat}
    \begin{tabular}{l cccc}
    \toprule
          \textbf{Model} & \textbf{Type} & MAE $\downarrow$    & RMSE $\downarrow$ & $R^2 \uparrow$ \\
         \midrule
        DRFP & ML & \underline{0.2741 $\pm$ 0.0018} & \textbf{0.4027 $\pm$ 0.0026} & \textbf{0.6701 $\pm$ 0.0043} \\
        MFF & ML & \textbf{0.2665 $\pm$ 0.0060} & \underline{0.4095 $\pm$ 0.0228} & \underline{0.6580 $\pm$ 0.0368} \\
        Chemprop & DL & 0.3053 $\pm$ 0.0136 & 0.4491 $\pm$ 0.0147 & 0.5893 $\pm$ 0.0268 \\
        ReaMVP & DL & 0.4307 $\pm$ 0.0104 & 0.6718 $\pm$ 0.0184 & 0.0813 $\pm$ 0.0512 \\
        T5Chem & DL & 0.5026 $\pm$ 0.0256 & 0.6441 $\pm$ 0.0175 & 0.1557 $\pm$ 0.0464 \\
         RXNFP & DL & 0.4276 $\pm$ 0.0137 & 0.6450 $\pm$ 0.0135 & 0.1534 $\pm$ 0.0356 \\
         Ours & DL & 0.3316 $\pm$ 0.0095 & 0.4477 $\pm$ 0.0117 & 0.5921 $\pm$ 0.0217 \\

         \midrule
        RXNFP* & DL & 0.3924 $\pm$ 0.0574 & 0.5372 $\pm$ 0.0562 & 0.4078 $\pm$ 0.1268 \\
         T5Chem* & DL & \underline{0.3274 $\pm$ 0.0178} & \underline{0.4734 $\pm$ 0.0334} & \underline{0.5424 $\pm$ 0.0650} \\
         ReaMVP* & DL & 0.3948 $\pm$ 0.0186 & 0.5039 $\pm$ 0.0226 & 0.4826 $\pm$ 0.0470 \\
         ours$^\dag$ & DL & \textbf{0.2699 $\pm$ 0.0092} & \textbf{0.3803 $\pm$ 0.0132} & \textbf{0.7055 $\pm$ 0.0203} \\
   \bottomrule
    \end{tabular}
\end{table}

\begin{table}[htbp]
    \centering
    
    \caption{\textbf{The results on chiral phosphoric acid-catalyzed thiol addition dataset(Sub test set).} The best performance of pretrained and not pretrained model is in \textbf{bold} and the second-best performance of pretrained and not pretrained models is underlined. * indicates that the model has been pretrained, and $\dag$ indicates that our model uses a pretrained condition encoder. }\label{tab: sel2_sub}
    \begin{tabular}{l c ccc}
    \toprule
    \textbf{Model} & Type & MAE $\downarrow$    & RMSE $\downarrow$ & $R^2 \uparrow$ \\
    \midrule
    DRFP & ML & 0.1873 $\pm$ 0.0066 & 0.2631 $\pm$ 0.0082 & 0.8551 $\pm$ 0.0091 \\
    MFF & ML &0.2741 $\pm$ 0.0061 & 0.3843 $\pm$ 0.0107 & 0.6910 $\pm$ 0.0169 \\
    Chemprop & DL & \underline{0.1487 $\pm$ 0.0098} & \underline{0.2036 $\pm$ 0.0105} & \underline{0.9132 $\pm$ 0.0091} \\
    ReaMVP & DL & 0.2223 $\pm$ 0.0875 & 0.2923 $\pm$ 0.1043 & 0.8032 $\pm$ 0.1517 \\
    T5Chem & DL & 0.4610 $\pm$ 0.0187 & 0.5885 $\pm$ 0.0211 & 0.2749 $\pm$ 0.0515 \\
    RXNFP & DL & 0.3783 $\pm$ 0.0229 & 0.4610 $\pm$ 0.0228 & 0.5547 $\pm$ 0.0446 \\
    Ours & DL & \textbf{0.1469 $\pm$ 0.0045} & \textbf{0.1971 $\pm$ 0.0039} & \textbf{0.9187 $\pm$ 0.0032} \\
    \midrule
    RXNFP* & DL & \underline{0.1849 $\pm$ 0.0407} & \underline{0.2381 $\pm$ 0.0458} & \underline{0.8779 $\pm$ 0.0503} \\
    T5Chem* & DL & 0.2173 $\pm$ 0.0262 & 0.2935 $\pm$ 0.0198 & 0.8192 $\pm$ 0.0247 \\
    ReaMVP* & DL & 0.3135 $\pm$ 0.0227 & 0.4002 $\pm$ 0.0313 & 0.6634 $\pm$ 0.0541 \\
    Ours$^\dag$ & DL & \textbf{0.1485 $\pm$ 0.0024} & \textbf{0.2009 $\pm$ 0.0033} & \textbf{0.9156 $\pm$ 0.0028} \\
   \bottomrule
    \end{tabular}
\end{table}

\begin{table}[htbp]
    \centering
    
    \caption{\textbf{The results on chiral phosphoric acid-catalyzed thiol addition dataset(Sub-Cat test set).} The best performance of pretrained and not pretrained model is in \textbf{bold} and the second-best performance of pretrained and not pretrained models is underlined. * indicates that the model has been pretrained, and $\dag$ indicates that our model uses a pretrained condition encoder. }\label{tab: sel2_cat_sub}
    \begin{tabular}{l c ccc}
    \toprule
          \textbf{Model} & Type & MAE $\downarrow$    & RMSE $\downarrow$ & $R^2 \uparrow$ \\
         \midrule
    DRFP & ML & \textbf{0.3213 $\pm$ 0.0080} & \underline{0.4846 $\pm$ 0.0103} & \underline{0.5643 $\pm$ 0.0186} \\
    MFF & ML & 0.3943 $\pm$ 0.0201 & 0.5616 $\pm$ 0.0300 & 0.4136 $\pm$ 0.0604 \\
    Chemprop & DL & \underline{0.3345 $\pm$ 0.0178} & \textbf{0.4747 $\pm$ 0.0191} & \textbf{0.5815 $\pm$ 0.0339} \\
    ReaMVP & DL & 0.5602 $\pm$ 0.1106 & 0.7824 $\pm$ 0.0958 & -0.1492 $\pm$ 0.2900 \\
    T5Chem & DL & 0.5521 $\pm$ 0.0125 & 0.7496 $\pm$ 0.0213 & -0.0429 $\pm$ 0.0592 \\
    RXNFP & DL & 0.4975 $\pm$ 0.0129 & 0.7109 $\pm$ 0.0179 & 0.0620 $\pm$ 0.0465 \\
    Ours & DL & 0.3581 $\pm$ 0.0126 & 0.4912 $\pm$ 0.0161 & 0.5521 $\pm$ 0.0297 \\
    \midrule
    RXNFP* & DL & \underline{0.3697 $\pm$ 0.0793} & 0.6053 $\pm$ 0.0641 & 0.3142 $\pm$ 0.1471 \\
    T5Chem* & DL & 0.3920 $\pm$ 0.0359 & \underline{0.5573 $\pm$ 0.0492} & \underline{0.4203 $\pm$ 0.1009} \\
    ReaMVP* & DL & 0.5403 $\pm$ 0.0367 & 0.7090 $\pm$ 0.0460 & 0.0644 $\pm$ 0.1225 \\
    Ours* & DL & \textbf{0.2977 $\pm$ 0.0090} & \textbf{0.4215 $\pm$ 0.0151} & \textbf{0.6702 $\pm$ 0.0234} \\
   \bottomrule
    \end{tabular}
\end{table}

\subsection{Qualitative Analysis of Attention Distributions}
\subsubsection{Scope, Limitations, and Caveats of this Qualitative Illustration}
We provide the following explicit caveats regarding the interpretation of the attention weights presented in this section. \textbf{It is crucial to emphasize that these visualizations are intended for illustrative purposes only and do not constitute causal explanations of the model's decision-making process.} The chemical narratives provided below are post-hoc interpretations that, in these specific instances, appear to align with chemical intuition. \textbf{These observations are intended to offer a qualitative glimpse into how the model organizes input information and should not be interpreted as systematic evidence of intrinsic chemical reasoning capabilities, nor as a substitute for rigorous quantitative interpretability metrics}. The following cases were selected to illustrate the range of attention patterns observed and do not represent a quantitative analysis of model behavior.
\subsubsection{Illustrative Case Studies: Observed Attention Patterns}
To qualitatively observe how the proposed Reaction-Center-Aware (RC-aware) cross-attention mechanism influences the model's focus, we visually compare the attention weight distributions of the RC-aware heads, the global heads within our full model, and the attention heads of a non-RC-aware baseline variant. This baseline is identical to our full model but employs a standard Transformer decoder with multi-head cross-attention over all encoder nodes, leaving the Atom-Aligned Encoder and all other components unchanged. All models are trained on the USPTO-Condition dataset. We present four representative cases below. The attention weights of each node in the reactants and products, while predicting the reaction condition combination, are visualized.
\begin{itemize}
\item \textbf{Case 1: Reduction of an Aromatic Nitro Group}
    
    Fig.~\ref{fig:case_study:specific components} presents a representative case of the reduction of an aromatic nitro group, with attention weights averaged across heads of each type. This transformation is typically achieved via metal catalyzed hydrogenation, a common industrial method valued for its relative eco friendliness and efficiency~\cite{2016Recent}. However, this approach can become unsuitable when the aromatic ring bears competing groups, such as halogens~\cite{2007207178.nh,HGJZ198404004}.
    
     In the non-RC-Aware baseline, the model consistently focuses on atoms within topologically complex structures (e.g., rings) or the reaction centers across all prediction steps, with attention weights distributed relatively evenly among all atoms of the molecule. This undifferentiated pattern offers limited insight into which specific structural features inform the model's decisions for different reaction components.
    
     In contrast, our RC‑Aware decoder exhibits a clear division of labor by design. The RC‑constrained heads are architecturally restricted to attend exclusively to the reaction center. This fixed focus on the transformation site anchors the model's understanding of the reaction type. Consequently, the remaining global heads are liberated from this responsibility and can flexibly attend to peripheral substituents with task‑specific patterns. During catalyst prediction, the global heads exhibit distinct attention on the presence of halogen atoms on the aromatic ring, a structural feature that can influence catalyst selection. During solvent prediction, they shift focus to regions outside the reaction center, aligning with the structural features relevant to solvent selection. In reagent prediction, while the RC‑constrained heads attend to the leaving group, which ultimately combines with hydrogen to complete the reduction, the global heads simultaneously monitor other relevant functional groups.

    \begin{figure}[htbp]
    \centering
    \includegraphics[width=0.9\linewidth]{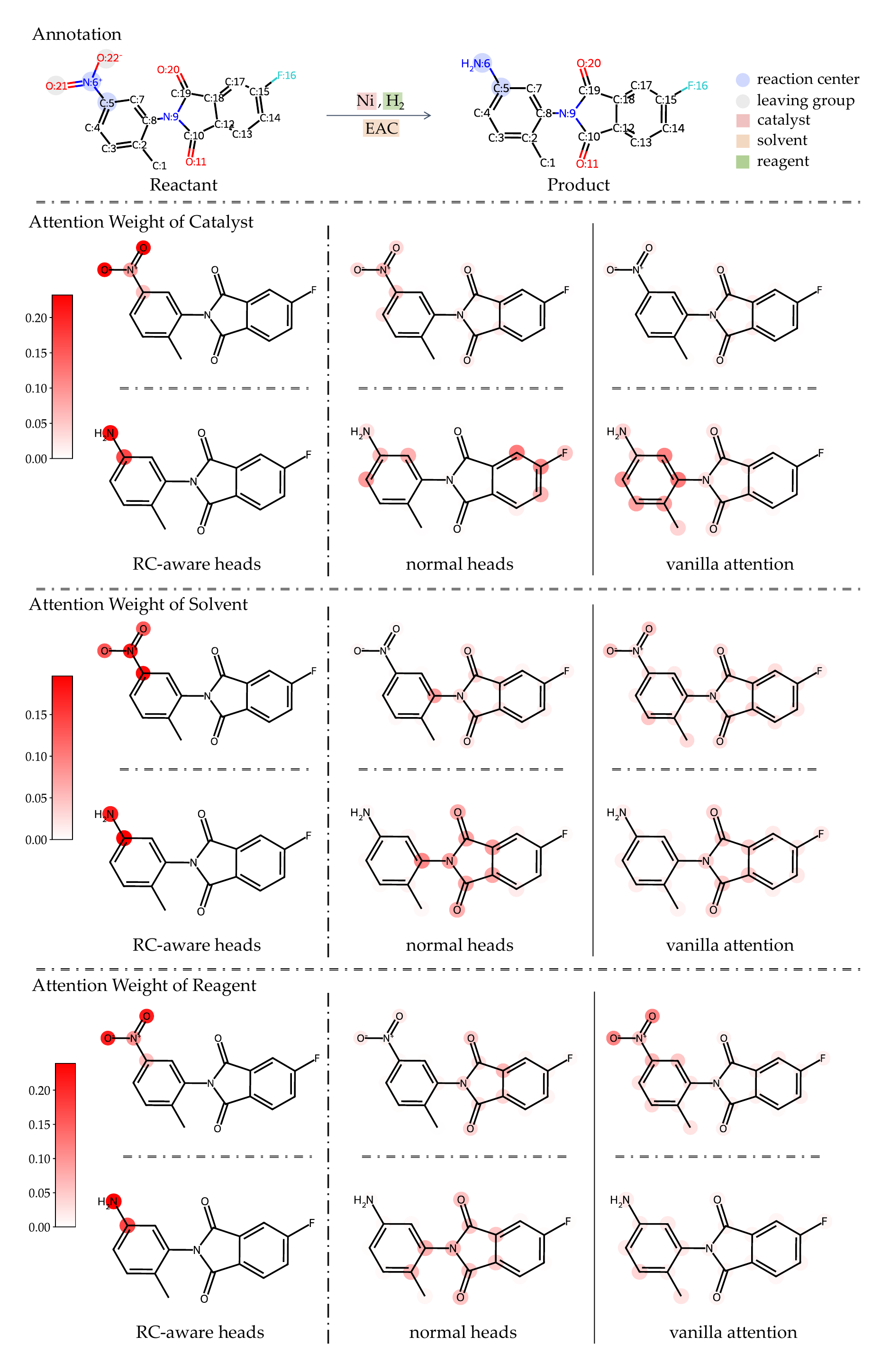}
    
    \caption{\textbf{The visualization the attention weights of each node in reactants and products while predicting the reaction condition combination for the reduction of nitro groups.}}
    \label{fig:case_study:specific components}
\end{figure}

\item \textbf{Case 2: Reduction of an Aldehyde}

In the second case presented in Supplementary Fig.~\ref{fig: supp_cass_1}, which involves the reduction of an aldehyde, the prediction of the reagent sodium borohydride draws the attention of the non-RC-Aware baseline to the reaction center, as well as to the topologically complex tert-butyl group and aromatic ring. However, it overlooks the ester group on the opposite side of the molecule, which represents a potential competing site for reduction and can be reduced by the more reactive reagent lithium aluminum hydride, included in the candidate reagent library of the USPTO-Condition dataset. In contrast, while the RC-Aware heads focus on the reaction center, the normal heads of our model concurrently attend to this potential competing site, reflecting a more comprehensive consideration of the molecular context.

\begin{figure}
    \centering
    \includegraphics[width=\linewidth]{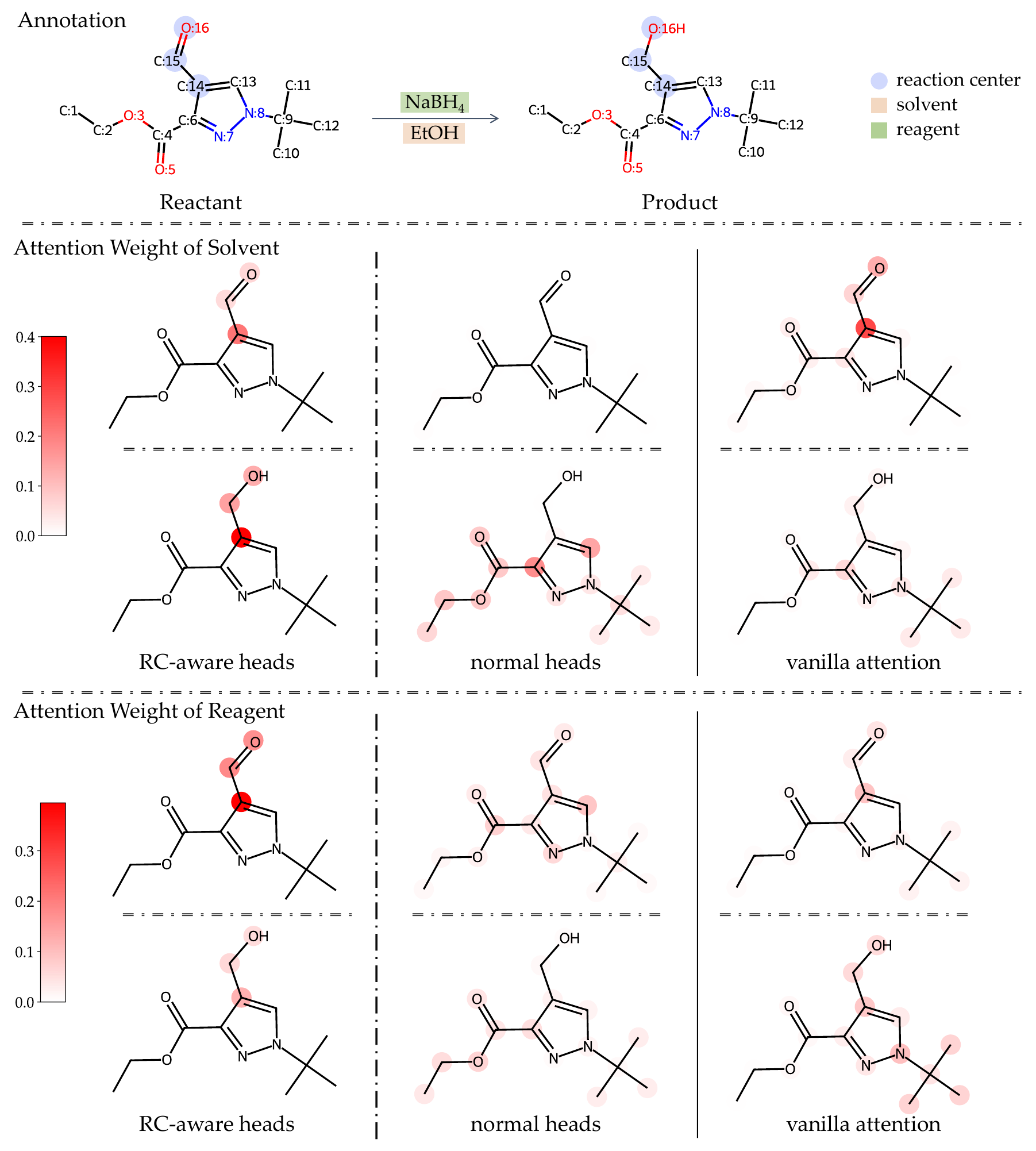}
    
    \caption{Supplementary case one on the RC-Aware cross attention mechanism. The attention weights of each node in reactants and products while predicting the reaction condition combination of the given reaction is visualized.}
    \label{fig: supp_cass_1}
\end{figure}
\item \textbf{Case 3: Synthesis of a Ketone from a Weinreb Amide}

In the reaction shown in Supplementary Fig.~\ref{fig: supp_cass_2}, a ketone is synthesized from a Weinreb amide and 4-methylpyridine using a lithium reagent. Lithium diisopropylamide (LDA) is selected as the base for this transformation because the reaction requires a metal ion, such as lithium, to form a stable five-membered chelated intermediate with the Weinreb amide. This chelation is essential for facilitating subsequent mild nucleophilic addition~\cite{Zhao_Wei55,nahm1981n}. Additionally, the strong electron-withdrawing nature of the pyridine ring enhances the acidity of the 4-methyl protons in 4-methylpyridine, enabling a strong base to deprotonate this site and generate the carbon nucleophile. The ideal reagent must, therefore, fulfill three roles: it must provide a metal ion for chelation catalysis, it must be sufficiently basic to deprotonate 4-methylpyridine, and it must exhibit low nucleophilicity to avoid direct attack on the Weinreb amide. LDA, as a sterically hindered lithium amide base, satisfies all three requirements. %Its lithium ion participates in chelate formation, its strong basicity enables deprotonation, and its bulky diisopropyl groups effectively suppress undesired nucleophilic addition.  
Within this chemical context, the normal attention heads of our model exhibit a focus pattern that aligns closely with the underlying logic of the transformation. Significant attention is allocated to atoms \texttt{N:18} and \texttt{O:17} of the Weinreb amide, precisely the atoms involved in forming the essential five-membered chelation ring. Concurrently, the model maintains substantial attention on the entire pyridine ring of 4-methylpyridine, consistent with the electron-withdrawing character that renders its 4-methyl group a potential reactive site. This attention distribution suggests that the model integrates multiple factors influencing the reaction outcome.
\begin{figure}
    \centering
    \includegraphics[width=\linewidth]{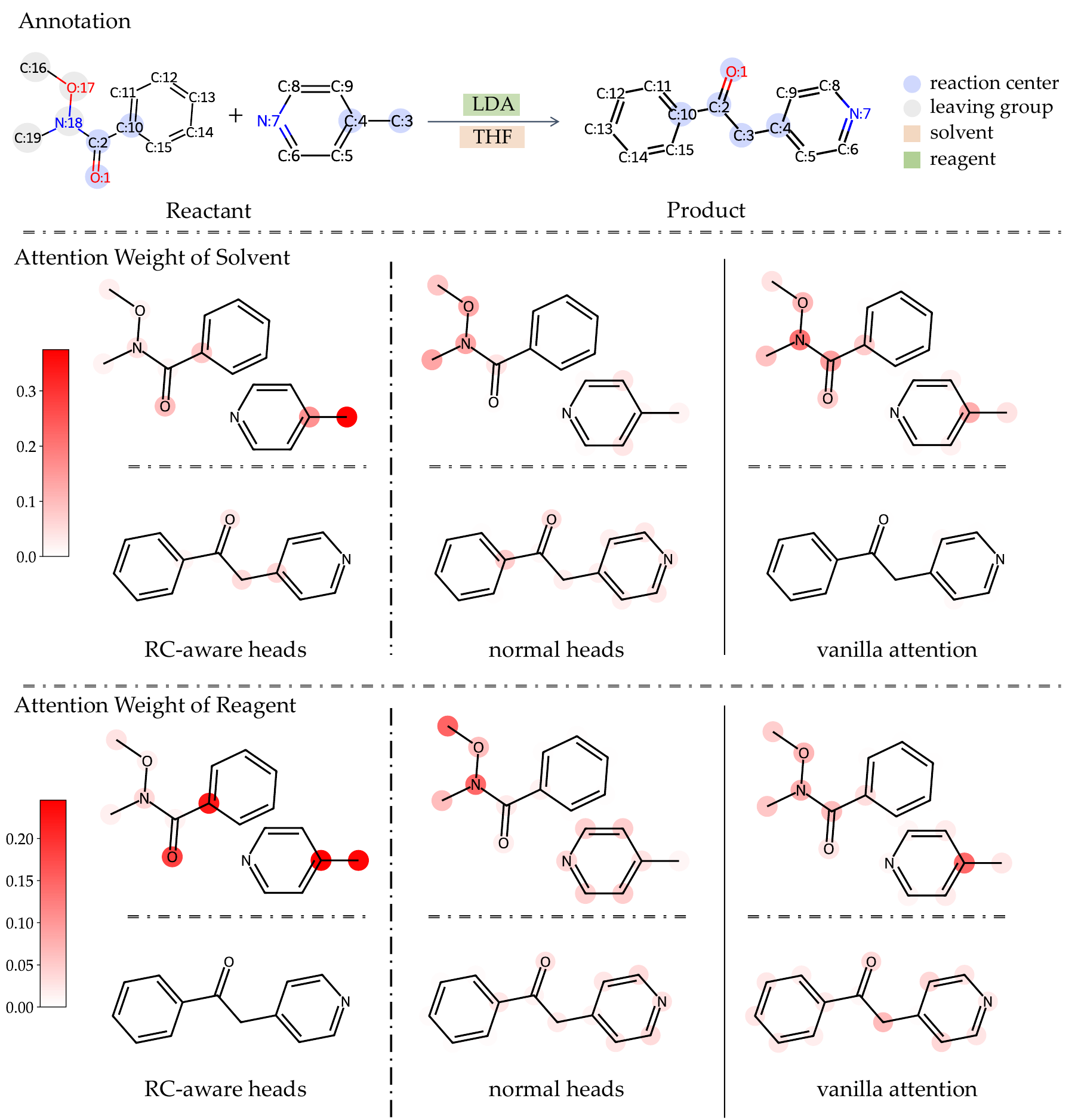}
    
    \caption{Supplementary case two on the RC-Aware cross attention mechanism. The attention weights of each node in reactants and products while predicting the reaction condition combination of the given reaction is visualized.}
    \label{fig: supp_cass_2}
\end{figure}

\item \textbf{Case 4: Selective Reduction of an Ester to an Aldehyde}

Supplementary Fig.~\ref{fig: supp_cass_3} shows that when predicting the reagent, the normal heads compensate for the attention of the RC-Aware heads to the aldehyde group at the reaction center in the product, which helps the model predict the correct reducing agent, diisobutylaluminum hydride, rather than a more powerful reductant, such as lithium aluminum hydride. 

\begin{figure}
    \centering
    \includegraphics[width=0.9\linewidth]{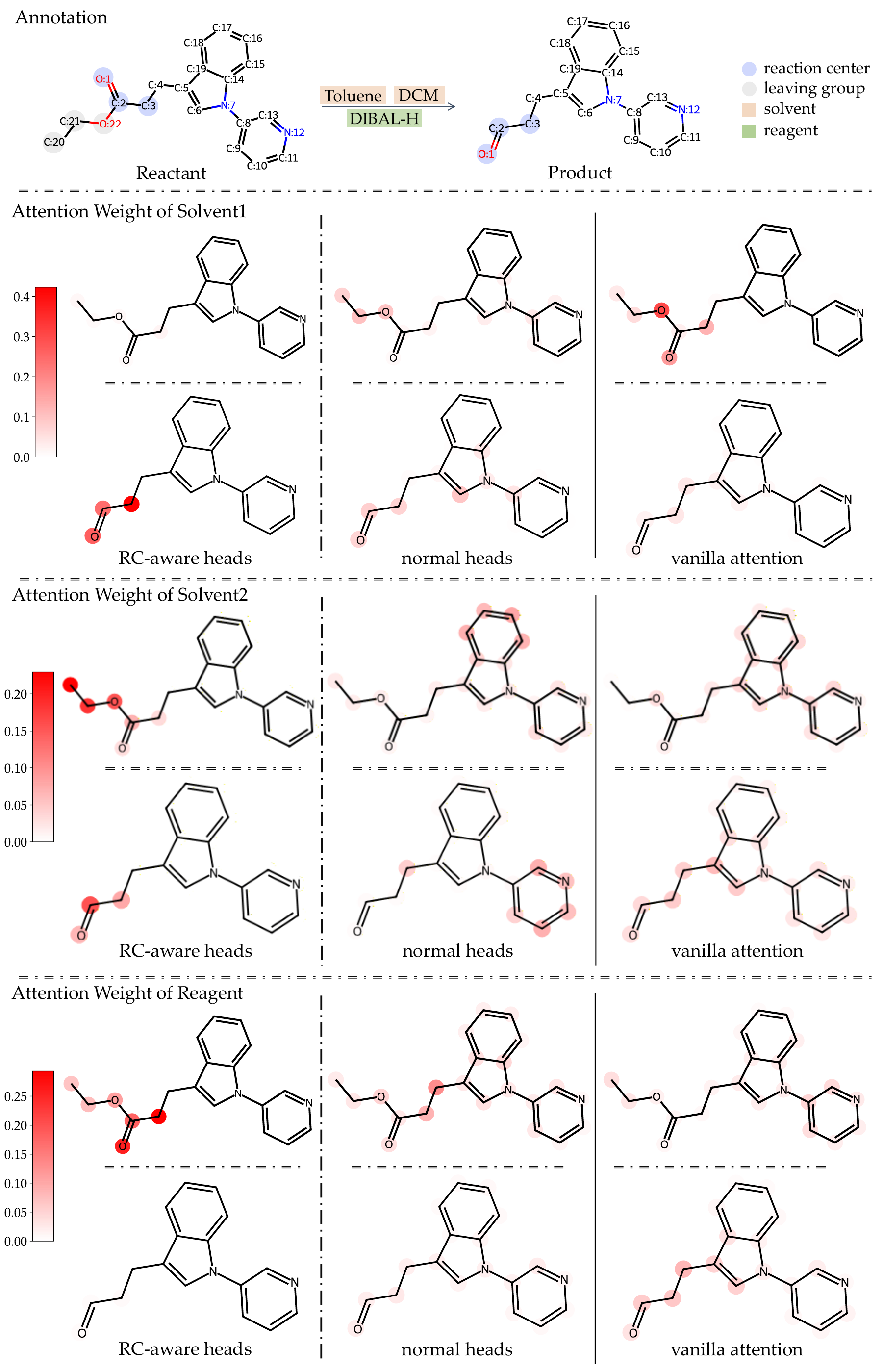}
    
    \caption{Supplementary case three on the RC-Aware cross attention mechanism. The attention weights of each node in reactants and products while predicting the reaction condition combination of the given reaction is visualized.}
    \label{fig: supp_cass_3}
\end{figure}

\end{itemize}
\subsubsection{Summary of Qualitative Observations}
Synthesizing the qualitative patterns observed across these four distinct reactions reveals several key characteristics of the RC-Aware framework's information aggregation. While conventional attention mechanisms (the non-RC-aware baseline) often concentrate focus on regions with high topological complexity or localized reaction centers, our framework facilitates a more structured partitioning of focus. Specifically, the results suggest a dual-stream attention strategy. The RC-aware heads, as architecturally intended, consistently focus on the predefined reaction center atoms. Meanwhile, the behavior of the global heads adapts to the chemical context: in complex molecules, they prioritize potential interfering functional groups or distal motifs relevant to condition selection. Notably, as illustrated in Cases 3 and 4, the global heads operate in synergy with the RC-aware heads, providing complementary reinforcement that stabilizes the representation of the reaction environment when distal functional groups are absent.

These observations do not constitute proof of the model's causal reasoning or its understanding of chemistry. However, the visual alignment between the observed attention patterns and chemically intuitive concepts in these specific instances suggests that the proposed architectural bias can lead to attention patterns that are more structured and context-dependent than the baseline. This observation presents a potential avenue for future hypothesis-driven research, which would require rigorous quantitative validation and perturbation-based analysis to establish any causal links.

\section{Implementation details} \label{impdetail}
\subsection{Initial node/edge feature extraction}
We use the atom and bond encoder provided by Open Graph Benchmark~\cite{hu2020open}. These encoders transform atoms and chemical bonds into integers based on their characteristics and molecular structures. Nine atom descriptors are provided, including the atom type, formal charge, and other properties that can be calculated by RDKit~\cite{landrum2013rdkit}. For chemical bonds, three descriptors are used: bond type, bond stereochemistry, and whether the bond is conjugated. Each descriptor corresponds to a learnable embedding table. The initial node and edge features are derived by aggregating the embeddings corresponding to each descriptor.

\subsection{MPNN Layer in the Atom-Aligned Encoder}
In contrast to many graphs where edges convey limited information, the chemical bonds within molecules play a pivotal role in determining their properties. Consequently, we employ a variant of the Graph Attention Network(GAT)~\cite{velivckovic2018gat}, akin to those utilized in the works of Yan et al.~\cite{yan2020retroxpert}  and Sacha et al.~\cite{megan}, to integrate information from chemical bonds into the node features during the message-passing process. The MPNN layer we implement can be mathematically formulated as follows:
\begin{equation}\label{eq: sgat}
\begin{aligned}
    &\tilde{e}_{u,v}  = \mathrm{FFN} _e (e_{u,v}  ),\\
    &\tilde{h}_{u}  = \mathrm{FFN} _n (h_u ),\\
    &c_{u,v} = \mathbf{a}^T [\tilde{h}_u  \Vert \tilde{h}_v \Vert \tilde{e}_{u,v} ],\\
    &\alpha_{u,v} = \frac{\exp(\mathrm{LeakyReLU}(c_{u,v}))}{\sum_{v'\in \mathcal{N}(u)\cup \{u\}} \exp(\mathrm{LeakyReLU}(c_{u,v'}))},\\
    &h'_u = \sum_{v\in \mathcal{N}(u) \cup \{u\}} \alpha_{u,v} \left(\tilde{h}_{u}  + \tilde{e}_{u,v} ^ {(k)}\right),\\
    \end{aligned}
\end{equation}
where $h_u$ represents the input node feature of node $u$, $e_{u,v}$ represents the input edge feature of edge $(u,v)$, and $h'_u$ represents the output node feature of $u$ in the message passing layer.
\subsection{Condition Encoders for Buchwald–Hartwig Dataset and thiol addition Selectivity Dataset}\label{app: condenc}
The Buchwald–Hartwig and thiol addition selectivity datasets employ chemical reagents as reaction conditions, with condition features generated via graph neural networks. Given the limited diversity of reagents in these datasets, and considering that several baselines were pretrained on large-scale reaction corpora, we adopted Hu et al.'s pretrained molecular representation model~\cite{Hu*2020Strategies} as our condition encoder for a fair comparison. This lightweight model underwent pretraining exclusively through random atom masking. The output representation of its $k$-th layer is formalized as:
\begin{equation}
\begin{aligned}
    &X = \sum_{u\in \mathcal{N}(v)\cup \{v\}} h^{(k-1)}_u + \sum_{e=(u,v)\:u\in \mathcal{N}(v)\cup \{v\}} h_e^{(k-1)},\\
    &h_v^{(k)} = {\rm ReLU}\left({\rm MLP}^{(k)}\left( X\right)\right),
\end{aligned}
\end{equation}
where $h_u^{(k)}$ is the node feature of node $u$ in the $k$-th layer, and $h^{(k)}_e$ is the edge feature of edge $e$ in the $k$-th layer. To conduct pure architectural comparisons, we also implemented a model using a simple GAT as the condition encoder, trained from scratch, and benchmarked it against the non-pretrained versions of all baselines. In these experiments, the simple GAT implementation follows Eq.~\ref{eq: sgat}.

\subsection{Model Implementation details}
We implement our model based on \texttt{PyTorch 1.13}~\cite{paszke2019pytorch} and \texttt{torch\_geometric 2.2.0}~\cite{Fey/etal/2025}.  All models are trained using the Adam optimizer~\cite{kingma2014adam}. 
We summarize the training and model architecture hyper-parameters in Supplementary Table~\ref{tab: cond_rec_para}, Supplementary Table~\ref{tab: para_regressions}, and Supplementary Table~\ref{tab: para_regressions2}. For the reaction condition recommendation task, we use the following strategy to adjust the learning rate during training: we gradually increase our learning rate to the maximum during the first few epochs and then decrease it using exponential decay. An early stopping strategy is also used, which halts the training when the evaluation metrics on the validation set do not increase for 20 epochs. 
\begin{table}[htbp]
    \centering
    
    \caption{The  training and model architecture hyper-parameters for reaction condition recommendation.}
    \begin{tabular}{|l| c | c|}
    \hline
        dataset & USPTO-Condition & USPTO-500MT \\
    \hline     
        hidden size & 384 & 256\\
    \hline
    learning rate& 1.5e-4 & 1.5e-4 \\
    \hline
    dropout ratio & 0.1 & 0.1 \\
    \hline
    number of encoder layers & 6 & 6\\
    \hline 
    number of decoder layers & 6 & 6\\
    \hline
    number of attention heads & 6 & 8 \\
    \hline
    number of training epochs & 200 & 200 \\
    \hline
    \end{tabular}
    \label{tab: cond_rec_para}
\end{table}

\begin{table*}[htbp]
    \centering 
    \caption{The  training and model architecture hyper-parameters for the regression tasks (with pretrained condition encoder). }
    \resizebox{\linewidth}{!}{
    \begin{tabular}{|l|c|c|c|}
    \hline
        dataset & Buchwald-Hartwig & radical C-H functionalization & thiol addition \\
        \hline
         hidden size of main model& 128  & 128 & 128\\
    \hline
    number of layers of main model & 3  & 5 & 3\\
    \hline 
    hidden size of condition encoder & 300  & - & 300 \\
    \hline
    number of layers of condition encoder & 5  & - & 5 \\
    \hline 
    learning rate & 5e-4  & 5e-4 & 5e-5\\
    \hline
    dropout ratio & 0.1  & 0 & 0\\
    \hline
    number of attention heads & 8  & 8 & 8 \\
    \hline
    number of training epochs & 200 & 200 & 250\\
    \hline
    \end{tabular}
    }
    \label{tab: para_regressions}
\end{table*}

\begin{sidewaystable}[htbp]
    \centering 
    \caption{The  training and model architecture hyper-parameters for the regression tasks (without pretrained condition encoder). }
    \begin{tabular}{|l|c|c|c|c|c|}
    \hline
        dataset & \multicolumn{2}{c|}{Buchwald-Hartwig} & radical C-H functionalization & thiol addition & USPTO-Yield \\
        \hline
        datasplit & random & OOD & - & all & - \\
         \hline
         hidden size of main model& 128 & 64 & 128 & 128 & 384\\
    \hline
    number of layers of main model & 3 & 3 & 5 & 3 & 6 \\
    \hline 
    hidden size of condition encoder & 128 & 64 & - & 128 & 384 \\
    \hline
    number of layers of condition encoder & 3 & 3 & - & 3  & 6 \\
    \hline 
    learning rate & 5e-4 & 5e-4 & 5e-4 & 5e-5 & 1e-4\\
    \hline
    dropout ratio & 0.1 & 0.1 & 0 & 0 & 0.1\\
    \hline
    number of attention heads & 8 & 8 & 8 & 8 & 8 \\
    \hline
    number of training epochs & 200 & 200 & 200 & 250 & 200\\
    \hline
    \end{tabular}
    \label{tab: para_regressions2}
\end{sidewaystable}

The training is conducted on Ubuntu 22.04.5, with two NVIDIA RTX 6000 GPUs. The training on the USPTO-Condition dataset uses two GPUs and takes around 60 hours. The training of other datasets is conducted on a single GPU. The training on the USPTO-500MT dataset takes around 24 hours with early stopping. The training for yield prediction on a single split with a single random seed takes around 2 hours. The training for selectivity prediction on the radical C-H functionalization dataset takes approximately 2 hours with a single split and a single random seed. The training on the chiral phosphoric acid-catalyzed thiol addition dataset with a single split and a single random seed takes around 1 hour.
\section{Baselines}\label{sec: baseline_detail}

\subsection{Chemical Reaction Condition Prediction}
We chose T5Chem~\cite{lu2022unified}, Parrot~\cite{Wang2023}, GCNN~\cite{maser2021multilabel}, Reagent Transformer~\cite{andronov2023reagent}, and FPRCR~\cite{Gao2018} as our baselines for chemical reaction condition recommendation; the details are as follows.
\begin{itemize}
    \item Parrot is a transformer-based model that accepts reaction SMILES as input to predict reaction condition combinations. It specifically targets condition sets with a fixed set of components. The model was pretrained on curated SMILES from USPTO 1976-2016sep~\cite{Lowe2017} using standard BERT-style masked language modeling~\cite{DBLP:conf/naacl/DevlinCLT19}.
    \item Reagent Transformer is a transformer-based model applied to both the USPTO-Condition and USPTO-500MT datasets. For a fair comparison on the USPTO-Condition dataset, we replaced its SMILES-generation vocabulary with a predefined molecular library and reformulated condition prediction as a fixed-length sequence generation task, adapting the original decoder design that produces SMILES strings.
    \item T5Chem is a language model pretrained on the PubMed dataset~\cite{10.1093/nar/gkaa971} using self-supervised learning and on the USPTO-500MT dataset~\cite{lu2022unified} with five supervised tasks, including reaction condition generation. We evaluated this baseline on both the USPTO-Condition and USPTO-500MT datasets. Analogous to our adaptation of Reagent Transformer for USPTO-Condition, we replaced T5Chem's SMILES-generation vocabulary with a predefined molecular library and reformulated condition prediction as a fixed-length sequence generation task producing five components.
    \item FPRCR is a fingerprint-based method specifically designed for reaction condition combinations with a fixed number of components. The model comprises a set of MLPs that correspond to the number of components in each reaction condition combination. All MLPs are invoked in the order of prediction, using the fingerprints of reactants, products, and the already predicted components of the reaction condition combination as input to forecast the next molecule that should be used as a condition.
    \item GCNN is a graph-based method that takes the molecular graphs of reactants and products as input to generate chemical reaction representations for predicting reaction condition combinations. The reaction condition combinations in the USPTO-Condition dataset consist of up to one catalyst, two solvents, and two reagents. We have extended the model's prediction head from a single multilabel classification head to five separate single-label classification heads.
\end{itemize}

\subsection{Chemical Reaction Yield/Selectivity Prediction}
The tasks of predicting chemical reaction yield and selectivity are highly analogous regression tasks; hence, their baselines are identical. \textbf{It is noteworthy that the emphasis of this paper is on introducing a novel architecture for chemical reaction representation learning, rather than being limited to the specific task of chemical reaction yield/selectivity prediction.} Therefore, we have not included baselines with complex pretraining tasks or intricate training loss and inference strategies. Pretraining and specialized loss functions tailored for reaction yield prediction are orthogonal to our research. The details of the chosen baselines are as follows.

 \begin{itemize}
     \item DRFP~\cite{probst2022reaction} is a type of chemical reaction fingerprint for yield prediction that calculates the fingerprint of the symmetric difference between the n-grams of reactants and products.
     \item MFF~\cite{MFF_SANDFORT20201379} utilizes multiple fingerprint features from all molecules in one reaction as input for the regressor.
     \item Chemprop~\cite{chemprop:doi:10.1021/acs.jcim.3c01250} is a type of Message Passing Network that encodes reactions. It converts the reactant and product SMILES pairs into a single pseudo-molecule known as the condensed graph representation (CGR) of the reaction and processes it through a directed messaging passing network (D-MPNN) to extract reaction features.
     \item  RXNFP~\cite{schwaller2021mapping} is a transformer model that has been pre-trained on the Pistachio dataset~\cite{mayfield2017pistachio} using self-supervised learning tasks. Building upon this work, the original research team has also proposed a finetuned version for predicting chemical reaction yields, termed YieldBert~\cite{schwaller2021prediction}.
     \item T5Chem~\cite{lu2022unified} is a language model pretrained on the PubMed dataset~\cite{10.1093/nar/gkaa971} with a self-supervised task and the USPTO-500MT dataset~\cite{lu2022unified} with five different supervised tasks, including reaction yield prediction.
     \item ReaMVP~\cite{shi2024prediction} is a multi-modal model that takes both SMILES and molecular graphs as input. It is pretrained on USPTO 1976-2016 sep~\cite{Lowe2017} and CJHIF~\cite{jiang2021smiles} using contrastive learning. 
     
 \end{itemize}

\section{Integrating Diverse Reaction Conditions: A Case Study}
To explore the extensibility of our proposed Condition Adapter and its potential to integrate non-molecular reaction conditions, we conducted a preliminary case study using the open-source USPTO 1976-2016sep dataset~\cite{Lowe2017} for yield prediction.

\subsection{Data Extraction}
We converted the original XML files to JSON and extracted the following information:
\begin{itemize}
    \item Reactant list and their corresponding amounts.
    \item Reagent list and their corresponding amounts.
    \item Product SMILES.
    \item Temperatures associated with all steps in the procedural sequence.
    \item Product yield.
    \item Mapped reaction SMILES.
\end{itemize}

\subsection{Data Curation Pipeline}
The extracted data underwent the following cleaning and standardization steps:
\begin{enumerate}
    \item \textbf{Atom Mapping Assignment:} Based on the mapped reaction SMILES, we assigned atom mapping numbers to each reactant and reagent. Components were matched by canonical SMILES, with priority given to reactants. Reactions with duplicate atom-mapping numbers were removed.
    \item \textbf{Re-refining Reactants \& Reagents:} Components containing atom mapping numbers present in the products were re-classified as reactants; all others were treated as reagents.
    \item \textbf{Amount Normalization:} We processed only three units: mole, mass, and volume. Other units were discarded (marked as \texttt{None}). Mass amounts were converted to moles using molecular weight. All amounts were normalized to base units: moles (mol) for quantity and liters (L) for volume. The smallest non-\texttt{None} molar amount among reactants was defined as the theoretical maximum yield (1.0 equivalent). All reagent amounts were normalized by this equivalent, and the reaction yield was calculated accordingly.
    \item \textbf{Temperature Assignment:} The temperature value that is furthest from room temperature among the recorded procedural steps was selected as the reaction temperature.
    \item \textbf{Data De-duplication:} Reactions were considered duplicates if they shared identical combinations of (a) reactants and their amounts, (b) products, (c) reagents and their amounts, and (d) reaction temperature. All duplicate entries were aggregated, and their yields were averaged to produce a single representative value for each unique reaction condition.
\end{enumerate}
Following this pipeline, we obtained 756,544  reactions, which were randomly split into training, validation, and test sets in an 8:1:1 ratio.

\subsection{Model Implementation and Experimental Results}
Reagents (molecular), their normalized amounts, and the reaction temperature were used as conditions. We trained two model variants:
\begin{itemize}
    \item Only molecular reagents were used as conditions.
    \item Molecular reagents, \textit{along with} their amounts and the reaction temperature, were used as conditions. To prevent label leakage, the amounts of the products were excluded.
\end{itemize}

\noindent\textbf{Implementation Details:}
\begin{itemize}
    \item Molecular reagents were encoded using a Simple GAT, whose message-passing layer follows Eq.~\ref{eq: sgat}.
    \item Numerical conditions (amounts, temperature) were projected into a $\mathbb{R}^{1 \times d}$ feature vector. We adopted a cluster-based embedding method~\cite{liu2024graph}, implemented as $\text{Linear}(\text{Softmax}(\text{Linear}(x)))$, where $x$ is the input numerical value.
    \item To integrate amount information into reagent embeddings, the adapter structure defined in Section 4.3 of the main text was used to fuse the outputs of the Simple GAT and the numerical condition embeddings.
\end{itemize}
\begin{table}[htbp]
    \centering
    
    \caption{Performance Comparison With and Without Non-Molecular Reaction Conditions}
    \begin{tabular}{ccccc}
    \toprule
       Amount & Temperature & $R^2\ \uparrow$ & MAE $\downarrow$ & RMSE $\downarrow$\\
         \midrule
        $\checkmark$ & $\checkmark$ & 0.2495 & 18.2864 & 22.7346\\
        $\checkmark$ & $\times$  & 0.2527 & 18.1232 & 22.6864\\
        $\times$ & $\checkmark$ & 0.2291 & 18.5609 & 23.0410\\
        $\times$ & $\times$ & 0.2373 & 18.2872 & 22.9180\\
    \bottomrule
    \end{tabular}
    \label{tab: num_abl}
\end{table}

\noindent The results are summarized in Supplementary Table~\ref{tab: num_abl}, which details the impact of incorporating two specific types of non-molecular conditions on the yield regression task. Interestingly, we observe that the inclusion of temperature as a condition leads to a decline in model performance across both model variants, regardless of whether the reagent amount is also included. We attribute this counterintuitive result to the inherent noise in the temperature data, which is parsed from the procedural sequence using a rule-based heuristic: the temperature value furthest from room temperature is selected as the reaction temperature. This heuristic introduces considerable uncertainty and may not accurately reflect the true reaction conditions. In contrast, when the reagent amount is introduced as an additional condition, a noticeable improvement in predictive accuracy is observed. This result suggests that despite the challenges posed by the noisy temperature feature, the integration of the reagent amount provides a meaningful and complementary signal that enhances model performance. Nevertheless, this experiment serves as a proof-of-concept demonstrating that our adapter architecture can technically accommodate diverse non-molecular condition modalities. While further refinement and higher-quality datasets are required to fully unlock its predictive benefits, it lays the groundwork for future developments in more comprehensive reaction representations.

\bibliography{sn-bibliography}
\end{document}